\documentclass[authoryear,preprint,3p,times,11pt]{elsarticle}

\usepackage{tikz}
\usepackage{txfonts}
\usepackage{natbib}
\usepackage{epsfig}
\usepackage{float}
\usepackage{graphicx}
\usepackage{epstopdf}
\usepackage{amssymb}
\usepackage[colorlinks,linkcolor=blue,hyperindex,CJKbookmarks]{hyperref}
\usepackage{bm}
\usepackage{amsmath}
\usepackage{pdflscape}
\usepackage{array}
\usepackage[english]{babel}
\usepackage[font=small,labelfont=bf,tableposition=top]{caption}
\usepackage{booktabs}
\usepackage{threeparttable}
\usepackage{lscape}
\usepackage{subfigure}
\usepackage{multirow}
\usepackage{upgreek}
\usepackage{comment}
\usepackage{setspace}
\usepackage{appendix}
\usepackage{mathtools}
\usepackage{cases}
\usepackage{xcolor}
\usepackage{hhline}
\usepackage{makecell}
\usepackage{soul}
\usepackage{ntheorem}
\usepackage{enumerate}
\usepackage{ulem}
\usepackage[displaymath,mathlines,edtable]{lineno}
\usepackage{lipsum}
\usepackage{cuted}
\usepackage[margins]{misc/trackchanges}
\usepackage{booktabs}
\usepackage{longtable}
\usepackage{array}
\usepackage{cleveref}

\usepackage{etoolbox} %% <- for \cspreto, \csappto
\usepackage[section]{placeins}

\usepackage[ruled]{algorithm2e} % 自己加入的伪代码库

\newcommand{\EQ}{\begin{eqnarray}}
\newcommand{\EN}{\end{eqnarray}}
\newcommand{\EQQ}{\begin{eqnarray*}}
\newcommand{\ENN}{\end{eqnarray*}}

\newcommand{\bremark}{\begin{remark} \begin{rm}}
\newcommand{\eremark}{ \end{rm} \rule{1mm}{2mm}
\end{remark}}
\newcommand{\basm}{\begin{assumption} \begin{rm}}
\newcommand{\easm}{\end{rm}
\end{assumption}}
\newcommand{\bsup}{\begin{supposition} \begin{rm}}
\newcommand{\esup}{\end{rm}
\end{supposition}}
\newcommand{\btheorem}{\begin{theorem} \begin{rm}}
\newcommand{\etheorem}{\end{rm} \rule{1mm}{2mm}
\end{theorem}}
\newcommand{\blemma}{\begin{lemma}\begin{rm}}
\newcommand{\elemma}{\end{rm} \rule{1mm}{2mm}
\end{lemma}}
\newcommand{\bcorollary}{\medskip\begin{corollary}\begin{rm}}
\newcommand{\ecorollary}{\end{rm}\rule{1mm}{2mm}
\end{corollary}}
\newcommand{\bdefinition}{\medskip\begin{definition}\begin{rm}}
\newcommand{\edefinition}{\end{rm}\rule{1mm}{2mm}
\end{definition}}
\newcommand{\bproposition}{\medskip\begin{proposition}\begin{rm}}
\newcommand{\eproposition}{\end{rm}\rule{1mm}{2mm}
\end{proposition}}
\newcommand{\bexample}{\begin{example}\begin{rm}}
\newcommand{\eexample}{\end{rm}\rule{1mm}{2mm}\end{example}}

\newcommand{\bcase}{\begin{case}\begin{rm}}
\newcommand{\ecase}{\end{rm}\rule{1mm}{2mm}\end{case}}

\newtheorem{theorem}{\bf Theorem}[section]
\newtheorem{lemma}{\bf Lemma}[section]
\newtheorem{definition}{\bf Definition}[section]
\newtheorem{remark}{\bf Remark}[section]
\newtheorem{corollary}{\bf Corollary}[section]
\newtheorem{proposition}{\bf Proposition}[section]
\newtheorem{example}{\bf Example}[section]
\newtheorem{assumption}{\bf Assumption}[section]

\newtheorem{case}{\bf Case}[section]
\newtheorem{supposition}{\bf Hypothesis}[section]

\def\equationautorefname~#1\null{(#1)\null}

\graphicspath{{Figure/}}% 添加图片路径

\journal{}

\begin{document}

\begin{frontmatter}
\title{PC-SNN: Predictive Coding-based Local Hebbian Plasticity Learning in Spiking Neural Networks}

\author[sysu]{Haidong Wang}
\ead{hdwang26@mail2.sysu.edu.cn}

\author[hit]{Xiaogang Xiong}
\ead{xiongxg@hit.edu.cn}

\author[hit]{Mengting Lan}
\ead{20s053097@stu.hit.edu.cn}

\author[cityu]{Yinghao Chu}
\ead{yinghchu@cityu.edu.hk}

\author[hit]{Zixuan Jiang}
\ead{190710110@stu.hit.edu.cn}

\author[usd]{KC Santosh}
\ead{kc.santosh@usd.edu}

\author[ln]{Shimin Wang}
\ead{smwang@ln.edu.hk}

\author[sysu]{Renxin Zhong\corref{cor1}}
\ead{zhrenxin@mail.sysu.edu.cn}

\cortext[cor1]{Corresponding author.}
\address[sysu]{School of Intelligent Systems Engineering,  Sun Yat-Sen University (Shenzhen Campus), Guangdong, China.}
\address[hit]{School of Mechanical Engineering and Automation, Harbin Institute of Technology Shenzhen, Guangdong, China.}
\address[cityu]{Department of Advanced Design and Systems Engineering, City University of Hong Kong, Hong Kong SAR.}
\address[usd]{Department of Computer Science, The University of South Dakota, SD, USA.}
\address[ln]{School of Data Science, Lingnan University, Tuen Mun, Hong Kong.}

\begin{abstract}
% Spiking Neural Networks (SNNs), regarded as the third generation of neural networks, emulate the brain’s information processing with unparalleled biological plausibility compared to traditional neural networks. 
{Spiking Neural Networks (SNNs), recognized as the third generation of neural networks, simulate the brain's information processing with a higher degree of biological realism than conventional neural networks.}
However, their non-linear, event-driven dynamics pose significant challenges for training, and existing methods often deviate from neuroscientific principles of cortical learning. Drawing inspiration from predictive coding theory—a leading model of brain information processing—we propose PC-SNN, a novel learning framework that integrates predictive coding with SNNs to enable biologically plausible, local Hebbian plasticity without reliance on backpropagation. Unlike conventional SNN training approaches, PC-SNN leverages only local computations, aligning with the brain’s distributed processing and overcoming the biological implausibility of global error propagation. Our classification model achieves competitive performance on the benchmark datasets, including Caltech Face/Motorbike, MNIST, NMNIST, and CIFAR-10. Furthermore, our predictive coding-based regression model outperforms backpropagation-based methods while adhering to local plasticity constraints, offering a scalable and biologically grounded alternative for SNN training. 
% PC-SNN not only advances neuromorphic computing by demonstrating the feasibility of direct implementation in spiking neural circuitry but also provides new insights into the neural mechanisms of learning, bridging neuroscience and computational models with a framework that is both theoretically innovative and practically effective.
PC-SNN drives progress in neuromorphic computing through validating the adaptability of bio-inspired algorithms within spiking neural architectures, but also unveils novel understandings of neurocognitive learning processes, presenting a conceptual framework distinguished by its theoretical originality and functional efficacy.
\end{abstract}


\end{frontmatter}
% \linenumbers

% \section{Introduction}

% \section{Methodology}
% \subsection{test.t}

% \appendix

%\input{abs.tex}
\section{Introduction}

{Spiking Neural Networks (SNNs), the third generation of neural networks, employ bio-inspired neurons that communicate by transmitting discrete binary spikes}, closely mimicking the event-driven processing of human biological neural systems \citep{review}. Renowned for their energy efficiency and biological plausibility, SNNs have demonstrated remarkable results on various tasks, particularly with event-driven data. These attributes make SNNs a promising paradigm for neuromorphic computing and neuroscience-inspired artificial intelligence. However, despite their potential, current SNN training methods often rely on techniques adapted from artificial neural networks (ANNs), which neglect the neuroscientific principles of cortical learning \citep{1999predictiveframework, lillicrap2020backpropagation, lillicrapRandomSynapticFeedback2016, caucheteuxEvidencePredictiveCoding2023}. This reliance introduces a significant gap, as such methods fail to fully exploit the biological fidelity of SNNs, limiting their scalability and alignment with neural mechanisms.

The non-differentiable nature of SNNs makes training challenging with popular and well-studied gradient descent methods. Here are three major approaches to train SNNs. 
First,
{the spike-timing-dependent plasticity (STDP) learning algorithm offers a biologically plausible mechanism for training SNNs by modulating synaptic weights based on the temporal relationship of pre- and post-synaptic spikes. However, the efficacy of the STDP approach may be constrained when applied to large-scale and computationally complex tasks \citep{stdp_mozafari2018first, kheradpisheh2018stdp}.}
Second, the ANN-SNN conversion approach aims to convert well-established ANN architectures to SNNs, benefiting from the high performance of ANNs and the energy efficiency of SNNs \citep{ann2snn_roy2019towards,ann2snn_han2020deep,ann2snn_jiang2023unified}. However, this approach has several drawbacks, including information loss and performance degradation.  
Third, the surrogate gradient-based direct learning approach addresses some limitations of the ANN-SNN conversion approach by enabling end-to-end training using surrogate gradients to approximate gradients during training \citep{sgsnn_guo2022real,sgsnn_zenke2018superspike}. This approach achieves competitive accuracy compared to other methods and offers the flexibility to train a wide range of network architectures.
However, the above studies on SNN training are all based on the backpropagation paradigm. Previous research has shown that feedback connections in the cerebral cortex do not exhibit the form of backpropagation \citep{1999predictiveframework, lillicrap2020backpropagation, lillicrapRandomSynapticFeedback2016}, and the nature of SNNs mimic the information encoding and processing of the human brain by using discrete events to simulate the propagation of pulse signals. Therefore, studying biological plausibility and brain-like learning mechanism in SNNs holds great significance \citep{aceituno2023learning, caucheteuxEvidencePredictiveCoding2023}.

Advancing SNNs requires moving beyond backpropagation, whose global error propagation and inter-layer communication conflict with both the distributed nature of neuromorphic hardware \citep{Dang2026, Goltz2025, Park2022} and the localized synaptic modifications observed in biological neural systems \citep{zipser1993neurobiological, Crick1989The}. These considerations motivate the development of local learning rules that rely solely on information accessible to individual synapses—the activities of directly connected neurons—thereby offering a pathway toward energy-efficient neuromorphic implementations while potentially illuminating the biochemical mechanisms underlying learning in the brain.
{The predictive coding theory, an influential theoretical framework in neuroscience,} exhibits valuable intriguing properties within the learning and perception in the brain \citep{1999predictiveframework, friston2003learning, friston2005theory, ororbia2024review, salvatori2023brain}. 
{Various methodologies have been formulated to approximate the backpropagation algorithm in non-spiking multilayer perceptron (MLP) networks, utilizing biologically plausible connectivity architectures and Hebbian learning principles \citep{2015Hebbian}.}  
{According to Whittington and Bogacz, the predictive coding framework—a hierarchical and biologically plausible process derived from a probabilistic model—is capable of yielding results comparable to those achieved by the backpropagation algorithm in ANNs \citep{whittington2017approximation,  1999predictiveframework, friston2003learning,friston2005theory}.}
{The predictive coding framework utilizes supplementary nodes to quantify localized prediction errors, which are defined as the discrepancy between the state of random variables at a specific hierarchical level and the predictions generated by subordinate layers. Furthermore, neural processes analogous to this propagation of prediction errors have been empirically documented within the context of perceptual decision-making tasks \citep{observe1, observe2}.}

We develop a predictive coding spiking neural network (PC-SNN) for image classification that leverages local Hebbian plasticity. Second, we implement a predictive-coding-through-time (PCTT) learning algorithm within an adaptive spiking recurrent neural network (ASRNN) for predicting angular velocities from event camera data.
Our results demonstrate excellent performance across these tasks. Compared to conventional SNN training methods, our Hebbian-inspired predictive coding approach is not only biologically plausible but also incorporates neurobiological constraints typically overlooked in artificial systems:
\begin{itemize}
  \item We proposed PC-SNN and ASRNN with PCTT, achieving competitive performance on classification and regression tasks on both traditional image and event camera datasets. 
  \item The predictive coding mechanism in SNNs achieves the basis for brain-like local computation and local plasticity, wherein neurons execute computations based solely on the activity of their input neurons and the synaptic weights associated with these inputs, rather than relying on information encoded elsewhere in the neural network. Additionally, synaptic plasticity is determined exclusively by the activity of presynaptic and postsynaptic neurons.
\end{itemize}

The remainder of this paper is organized as follows. Section 2 reviews related work on SNN training methods and biological plausibility to motivate the adoption of predictive coding and local Hebbian learning in later sections. Section 3 introduces preliminaries including temporal coding and neuron models of SNN. Section 4 presents our proposed methods: PC-SNN for classification tasks (Section 4.1); and regression tasks (Section 4.2), which includes the ASRNN model (Section 4.2.1), the ASRNN-BPTT algorithm as a baseline  (Section 4.2.2), and the ASRNN-PCTT algorithm that achieves local plasticity by replacing backpropagation with predictive coding (Section 4.2.3). Section 5 presents comprehensive experimental results demonstrating the effectiveness of both PC-SNN and ASRNN-PCTT, and Section 6 concludes with discussion. 

\section{Related Work}

\subsection{Spiking Neural Network training Methods}
Compared to the rapid development of ANNs, SNNs training  algorithms remain an active area of research due to their inherently non-differentiable nature. 
In general, SNN training algorithms fall into three categories: STDP  based localized learning rules \citep{stdp_bi1998stdp,stdp_mozafari2018first, kheradpisheh2018stdp,liu2021sstdp}, ANN-SNN conversion methodologies \citep{ann2snn_roy2019towards,ann2snn_han2020deep,ann2snn_jiang2023unified}, and surrogate gradient-based direct learning approach \citep{sgsnn_guo2022real,sgsnn_zenke2018superspike}. 
Unsupervised and semi-supervised learning algorithms based on STDP are limited to shallow SNNs,  often underperforming compared to ANNs on complex datasets.
The top-performing SNNs, usually made of Integrate-and-Fire (IF) neurons, are created through ANN-SNN transformations by training non-spiking ANNs with ReLU activation functions. 
% To deepen these networks, spike-based error backpropagation algorithms have been proposed for supervised SNN training.
{To facilitate the training of Spiking Neural Networks with greater architectural depth, researchers have developed spike-based error backpropagation algorithms specifically for supervised learning.}

% \cite{firingtime2} proposed a feedforward spiking network that implements a temporal coding scheme using spike times as the information preserver instead of sparse spike counts. 
{\cite{firingtime2} introduced a feedforward spiking network that uses a temporal coding scheme, where information is encoded directly in the precise timing of spikes rather than just in sparse spike counts.}
% They developed a direct training approach that does not require spiking networks to degenerate to conventional ANNs. By establishing a linear and differentiable relation between any given spike time and all spikes that have an impact on it in the z-domain, any differentiable cost function related to network spike times can be imposed through gradient descent verification. 
{A direct training methodology is presented that circumvents the need to convert spiking networks into conventional artificial neural networks. This approach is predicated on establishing a linear and differentiable relationship in the z-domain between a given spike event and all antecedent spikes that exert an influence upon it. This formulation enables the optimization of any differentiable, spike-time-dependent cost function through the application of a gradient descent algorithm.}
\cite{kheradpisheh2020S4NN} proposed a novel supervised learning algorithm named S4NN for multi-layer SNNs. It encodes input spikes such that spiking latency is inversely proportional to pixel value, enabling rapid and accurate decision-making with minimal spikes. Using rank-order coding, each neuron fires at most one spike per stimulus. Additionally, S4NN introduces relative target firing time to ensure correct output neuron activation and approximates ReLU using IF neurons.
\cite{Sun2024DelayLB} presented a Delay Learning based on Temporal Coding (DLTC) framework that jointly optimizes synaptic weights, axonal delays, and neuron thresholds within feedforward SNNs. Their approach demonstrated that learnable delays and thresholds enhance temporal precision, sparsity, and energy efficiency without impairing convergence, bridging temporal coding and surrogate-gradient learning for robust end-to-end training. In parallel, \cite{Shaban2021AnAT} introduced the Double Exponential Adaptive Threshold (DEXAT) neuron within recurrent SNNs to model homeostatic spike-frequency adaptation in hardware. Their results showed that incorporating adaptive thresholds can stabilize network dynamics and accelerate training convergence on sequential tasks while remaining robust to device variability when implemented on OxRAM-based neuromorphic circuits. 
Beyond the foundational temporal coding approaches, numerous studies have advanced Time-To-First-Spike (TTFS) based learning in SNNs. \cite{Zhang2020RectifiedLP} proposed the Rectified Linear Postsynaptic Potential (ReL-PSP) function, which addresses the gradient computation challenges in deep SNNs by providing a more tractable surrogate for temporal credit assignment. \cite{Zhou2019TemporalCodedDS} presented a temporal-coded deep SNN framework that achieves easy training through a novel temporal coding scheme combined with a robust loss function. More recently, \cite{Wei2023TemporalCodedSN} introduced a dynamic firing threshold mechanism for temporal-coded SNNs, enabling event-driven backpropagation that adapts to input statistics and improves learning efficiency.
These TTFS-based methods share a common principle of encoding information in precise spike timing rather than firing rates, which offers advantages in terms of energy efficiency and biological plausibility.

\subsection{Biological plausibility}
Spike-based error backpropagation algorithms adjust the strength of synapses/weights to minimize errors by backpropagating them through feedback networks. 
% However, \cite{zipser1993neurobiological} questioned whether a similar adaptive mechanism exists in biological neural systems. 
% Neuroscientists generally agree that there is no evidence of backpropagation occurring in biological nervous systems, primarily because such systems lack the specialized feedback networks that are integral to backpropagation algorithms but not found in biological neural systems \citep{Crick1989The}.
{Nonetheless, the biological plausibility of a corresponding adaptive mechanism within neural systems has been called into question \citep{zipser1993neurobiological}. A consensus exists within the neuroscience community that empirical evidence for backpropagation is absent in biological nervous systems. This conclusion is principally attributed to the lack of specialized feedback networks within biological systems, which are a fundamental prerequisite for the operation of backpropagation algorithms \citep{Crick1989The}.}
% Additionally, it is challenging to establish a one-to-one correspondence between synapses in feedforward and feedback networks or to adapt the strengths of corresponding connections during learning, as most connections between neurons involve multiple synapses.
{Furthermore, a significant challenge arises from the neuroanatomical fact that most inter-neuronal connections are composed of multiple synapses. This structural reality makes it problematic to establish a precise one-to-one correspondence between synapses in feedforward and feedback pathways, which consequently complicates the symmetrical adjustment of their respective weights during the learning process.}
% \cite{LinFeng} proposed a feedback-network-free algorithm for SNNs that is both essentially and mathematically equivalent to the traditional backpropagation algorithm.
{An algorithm for SNNs that operates without requiring a feedback network has been put forth by \citep{LinFeng}. This proposed method is demonstrated to be both conceptually and mathematically equivalent to the conventional backpropagation algorithm.}
This novel approach is inspired by retrograde regulatory mechanisms believed to exist in neurons and eliminates the need for a feedback network, thereby significantly enhancing the biological plausibility of learning capabilities in neural networks.

% Several models have been introduced in ANN learning algorithms that aim to implement backpropagation in multi-layer perceptron MLP-style models using only biologically plausible connectivity schemes and Hebbian learning rules. \cite{whittington2017approximation} demonstrated that the backpropagation algorithm can be closely approximated by employing a simple local Hebbian plasticity rule, which involves only the activities of presynaptic and postsynaptic neurons directly connected to the synapse being modified during learning. 
{Within the field of ANN learning algorithms, various models have been proposed to implement backpropagation in architectures similar to Multi-Layer Perceptrons (MLPs). These approaches are constrained to using only biologically plausible connectivity schemes and Hebbian learning principles. Notably, \cite{whittington2017approximation} demonstrated that the backpropagation algorithm can be approximated with high fidelity by applying a simple, local Hebbian plasticity rule. The modification of a given synapse under this rule depends exclusively on the activity of the directly connected presynaptic and postsynaptic neurons during the learning phase.}
Inspired by the predictive coding framework \citep{1999predictiveframework,friston2003learning,friston2005theory}, 
% their proposed model suggests that this form of inference could be implemented by biological neural networks.
{ the proposed model indicates that this form of inference is mechanistically plausible and could be realized within biological neural networks.}

% \cite{millidge2020predictive} recently demonstrated a strong correlation between predictive coding theory and automatic differentiation across arbitrary computation graphs, achieving the desired outcome on three mainstream machine learning architectures (CNNs, RNNs, and LSTMs). They utilized a predictive coding framework with local learning rules and mostly Hebbian plasticity. 
{A significant correspondence between predictive coding theory and automatic differentiation across arbitrary computational graphs was recently established by \cite{millidge2020predictive}. Their methodology, rooted in a predictive coding framework that utilizes local learning rules and predominantly Hebbian plasticity, was successfully validated on three widely-used machine learning architectures: CNNs, RNNs, and LSTMs.}
\cite{ororbia2023convolutional} applied predictive coding to convolution-based computations using 
% a flexible neurobiologically inspired algorithm that iteratively refines latent state feature maps to construct accurate internal representations of images. 
{an adaptive algorithm, informed by principles of neurobiology, that functions through the iterative refinement of latent state feature maps. The objective of this process is to construct veridical internal representations of visual images.}
Several investigations in the domain of convolutional-based SNNs have shown exceptional results, including ANN-SNN, Surrogate Gradient, Direct Training among others attaining state-of-the-art performance on challenging datasets such as CIFAR and ImageNet \citep{han2020rmp,li2021free,yao2023attention,li2022neuromorphic}.
\cite{song2020can} presented a brain learning model that achieves local plasticity and full autonomy through the use of a fully autonomous Z-IL (Fa-Z-IL) model. This model is equivalent to BP but does not require any control signal, allowing for simultaneous and autonomous computation at the local level. However, there is currently no formulation available for a biologically plausible predictive coding SNN with local Hebbian synaptic plasticity that draws inspiration from neuroscience.
\section{Preliminaries}

In this section, we introduce the fundamental concepts of our spiking neural network framework, including first-spike temporal coding, spiking neuron model, dynamic target firing time, and backpropagation-based training. These concepts form the theoretical foundation for our proposed predictive coding framework.

\subsection{First-spike temporal coding}
%Unlike the rate coding where the analog output of an ANN neuron is encoded in the spiking rates of a spiking neuron, we apply a sparser temporal coding strategy with the information embodied in the firing time. Specifically, neurons can only be activated once at most in a predefined time interval and neurons fire first are thought to be the most active.
%
%In the procedure of encoding the input image to spike trains, denote the $i^{th}$ pixel value of the input grayscale images as $P_{i}$, and the pixel values range in $\left[0,P_{\max}\right]$.  The larger the pixel value, the earlier the firing time. In order to inversely convert the pixel value to firing time in the range $\left[0,t_{\max}\right]$, we use the following linear transformation formula:
% We use a sparser temporal coding strategy that encodes the firing time of neurons, 
{The methodology employs a sparse temporal coding scheme wherein information is encoded via the precise firing times of neurons,}
rather than the analog output of an ANN neuron through spiking rates. 
% Specifically, 
{Specifically,each neuron is constrained to a single spike event within a predefined temporal window. The neuron's activation level is thereby encoded in its firing latency, with earlier spikes signifying a higher degree of activity.}
To encode input images into spike trains, we denote the $i^{th}$ pixel value as $P_{i}$ in grayscale images with pixel values ranging from $\left[0,P_{\max}\right]$. The larger the pixel value, the earlier it fires. To convert pixel values to firing times within $\left[0,t_{\max}\right]$, we use this linear transformation formula:
\begin{equation}\label{coding}
  t_{i}=\frac{P_{\max}-P_{i}}{P_{\max}} t_{\max}.
\end{equation}
% Subsequently, the input spike trains of the $i^{th}$ pixel (denote input layer as layer 0) can be represented as follows:
{Accordingly, for the $i^{th}$ pixel in the input layer (designated as Layer 0), the corresponding input spike train can be formulated as follows:}
\begin{equation}\label{input spike}
  S_{i}^{0}(t)= \begin{cases}1 & \text { if } t=t_{i} \\ 0 & \text { otherwise }\end{cases}
\end{equation}
Neurons in the hidden layers and output layer integrate incoming spikes over time with no leakage, which fire a solitary spike when their membrane potential first exceeds the threshold. 
% If a neuron's potential never reaches the threshold, it remains silent. During training, it is necessary to determine the firing time of all neurons. 
{A neuron remains in a quiescent state should its membrane potential fail to attain the requisite activation threshold. Accordingly, the training protocol necessitates the determination of the precise firing time for each neuron.}
% Therefore, if a neuron fails to reach threshold during training, we artificially set its activation time to $t_{max}$.
{Therefore, for any neuron that fails to achieve the firing threshold during the training phase, its activation time is assigned the maximum value, $t_{max}$.} 
%Indeed, neurons on the following hidden layers and output layer simply integrate weighted incoming spikes over time without leak and emit only one spike right after their membrane potentials cross the threshold for the first time, or are silent if their potentials never reach the threshold. Given that we need to know the firing time of all neurons during the training (see Eq.\eqref{a} and \eqref{eq:firingtimeupdate}), if a neuron does not reach the threshold during training, we artificially set its activation time to $t_{max}$.
%At the end of each training epoch, synapses connected to the neurons that were silent were randomly reweighted to a uniform distribution. For the output layer, each neuron is assigned to a category, and the category corresponding to the earliest neuron to fire is the final prediction of the network.

\subsection{Spiking neurons model}
%We use a non-leaky way to integrate neurons' potential and stimulate neurons that accumulate receieved spikes from presynaptic neurons through dendritic connection. Each presynaptic spike in the integration of its corresponding synaptic weight raises the neuron's potential. Emission occurs when their internal potentials hit a predefined threshold. The membrane potential of the $j^{th}$ neuron in the $l^{th}$ layer is described by:
We integrate neurons' potentials and activate them without leakage by accumulating received spikes that propagate from presynaptic neurons to postsynaptic neurons via their dendritic connections. Each presynaptic spike corresponds to a synaptic weight that raises the neuron's potential during integration. Neurons emit signals when their internal potentials reach a predefined threshold. The membrane potential of the $j^{th}$ neuron in the $l^{th}$ layer is described as follows:
\begin{equation}\label{IF}
  V_{j}^{l}(t)=\sum_{i} w_{j i}^{l} \sum_{\tau=1}^{t} S_{i}^{l-1}(\tau)
\end{equation}
% where $w_{j i}^{l}$ signifies the synaptic weight between the $i^{th}$ presynaptic neuron and the $j^{th}$ neuron in $l^{th}$ layer; and $S_{i}^{l-1}(\tau)$ denotes the spike train of the $i^{th}$ presynaptic neuron in layer $l-1$.
where $w_{j i}^{l}$ is the synaptic weight from the $i^{th}$ presynaptic neuron to the $j^{th}$ neuron in the $l^{th}$ layer, and $S_{i}^{l-1}(\tau)$ is the spike train of the $i^{th}$ presynaptic neuron in layer $l-1$.

If $V_{j}^{l}(t)$ transcends its threshold, then the neuron emits a spike:
\begin{equation}\label{spike}
  S_{j}^{l}(t)=\delta\left(t-t_{j}^{l}\right)= \begin{cases}1 & \text { if } V_{j}^{l}(t) \geq \theta_{j}^{l}  \wedge  S_{j}^{l}(<t) \neq 1 \\ 0 & \text { otherwise }\end{cases}
\end{equation}
where $S_{j}^{l}(<t) \neq 1$ means that the neurons do not fire until time $t$.

\subsection{Dynamic target firing time}\label{target firing time}
% One of the common methods for setting the target firing time is using fixed firing time: assuming that we have $C$ categories of input images and the label of one input image is $i$, then the target firing time of the $i^{th}$ neuron in the output layer is set to $T_{i}^{o}=\tau$, where $\tau$ is a given constant. 
{A prevalent approach for establishing target firing times is the fixed-time assignment method. Within a classification framework comprising $C$ categories, the output neuron corresponding to the input's ground-truth label, $i$, is assigned a target firing time of $T_{i}^{o}=\tau$, where $\tau$ is a predefined constant.}
% The activation time of the remaining output neurons is set to $T_{j}^{o}=t_{\max }, 1 \leq j \leq C, j \neq i$. Although the above setting method is simple and easy to implement, when the actual firing time of the $i^{th}$ neuron $t_{i}^{o}<\tau$, this setting affects the response speed of the network and can not meet our expectation of having the network respond as quickly as possible.
{The activation times for the remaining output neurons ($1 \leq j \leq C, j \neq i$) are designated as $T_{j}^{o}=t_{\max }$. While this methodology is characterized by its simplicity and ease of implementation, it can impose a suboptimal constraint on the network's response latency. Specifically, in instances where the actual firing time of the correct neuron ($t_{i}^{o}$) precedes the target value ($\tau$), this approach conflicts with the objective of minimizing the network's response time.}

% In this paper, we followed the dynamic target firing time proposed by S4NN model \citep{kheradpisheh2020S4NN} which takes the actual activation time into account.
This study adopts the dynamic target firing time methodology proposed by the S4NN model \citep{kheradpisheh2020S4NN}, a technique that incorporates the neuron's actual activation time.
Assuming that we feed the image of $i^{th}$ category to the network, the first step is to obtain the actual minimum activation time of the output neuron: $T=\min \left\{t_{j}^{o} \mid 1 \leq j \leq C\right\}$, then the dynamic target firing time of the $j^{th}$ output neuron is set as follows:
\begin{equation}\label{target}
  T_{j}^{o}=\left\{\begin{array}{lll}
T & \text { if } j=i & \\
\max\{T+\gamma, t_{j}^{o}\} & \text { if } j \neq i
\end{array}\right.
\end{equation}
where $\gamma>0$ is a constant. To clarify the choice of $T$ in the dynamic target firing formulation, we define $T$ as the minimum actual firing time among all output neurons for the current sample, ensuring that the correct class neuron is always encouraged to fire earlier than competing neurons. This adaptive, data-dependent definition prevents imposing an artificially fixed target time, maintains stable temporal ordering for predictive-coding inference, and is consistent with practices widely used in TTFS-based SNN models.

% In the above setting of dynamic target firing time, the output neuron of the true category is set to the minimum firing time. The positive constant $\gamma$ ensures the minimum distance between the firing time of the true category neuron and that of the remaining neurons. Neurons whose activation times are much later than $\tau$ do not change their values.
Within this dynamic target firing time framework, the output neuron corresponding to the ground-truth category is assigned the earliest firing time. A positive constant, $\gamma$, is utilized to enforce a minimum temporal separation between the spike time of the correct neuron and that of all other neurons. Furthermore, the target times for neurons with activation times that substantially exceed $T$ remain unaltered.

\subsection{SNN trained with backpropagation}
\par
% To demonstrate the theoretical advantages of PC-SNN over BP-SNN in terms of bio-plausibility, we will briefly review the SNN temporal-coding-based backpropagation algorithm in this section. For a $C$-category classification task, we define a temporal mean square error function as follows:
{To establish the theoretical advantages of PC-SNN over BP-SNN with respect to bio-plausibility, this section provides a concise examination of the backpropagation algorithm founded on temporal coding within Spiking Neural Networks. For a classification task comprising $C$ categories, the temporal mean square error function is formulated as follows:}

\begin{equation}
  L=\frac{1}{2} \sum_{j=1}^{C}\left(t_{j}^{o}-T_{j}^{o}\right)^{2}
\end{equation}
where $t_{j}^{o}$ and $T_{j}^{o}$ denote the actual and targeted firing time of the $j^{th}$ output neuron, respectively. 
% Let's define an intermediate gradient as $\delta_{j}^{l}=\frac{\partial L}{\partial t_{j}^{l}}$. Therefore backpropagation (BP) updates the weights of the SNN by:
{An intermediate gradient is formulated as $\delta_{j}^{l}=\frac{\partial L}{\partial t_{j}^{l}}$. Consequently, the backpropagation (BP) algorithm updates the synaptic weights of the SNN according to the following procedure:}
\begin{equation}\label{bp}
  \Delta w_{j i}^{l}\!=\!-\!\eta \frac{\partial L}{\partial w_{j i}^{l}}\!=\!-\!\eta \frac{\partial L}{\partial t_{j}^{l}} \frac{\partial t_{j}^{l}}{\partial w_{j i}^{l}}\!=\!-\eta \delta_{j}^{l} \sum_{\tau=1}^{t_{j}^{l}} S_{i}^{l-1}(\tau)
\end{equation}
where $\frac{\partial t_{j}^{l}}{\partial w_{j i}^{l}}= \sum_{\tau=1}^{t_{j}^{l}} S_{i}^{l-1}(\tau)$ is formulated detailedly in \citep{kheradpisheh2020S4NN}. The intermediate signal is given as follows:
\begin{equation}
\delta_{j}^{l}\!=\!\left\{\begin{array}{cc}
t_{i}^{o}-T_{i}^{o} & \text { if } l\!=\!o \\
\sum_{k} \delta_{k}^{l+1} w_{k j}^{l+1}\left[t_{j}^{l} \leq t_{k}^{l+1}\right] & \text { if } l \!\in\!\left\{1, \ldots, l_{\max }\!-\!1\right\}  
\end{array}\right.
\end{equation}
where $ l\!=\!l_{\max}$ when $l=o$.
%An example of two-layer SNN trained with BP is illustrated in Fig.~\ref{fig:SNN}(A). Note that the construction of a feedback structure is an essential step to backpropagate the intermediate gradient signal from $\delta_{k}^{l+1}$ to $\delta_{j}^{l}$, indicating that the weights of the feedback network must be brought into correspondence with the feed-forward network.
% Figure~\ref{fig:SNNvsPCSNN}(A) shows an example of a two-layer SNN trained with BP (BP-SNN).
An example of a two-layer BP-trained SNN (BP-SNN) is shown in Fig.~\ref{fig:SNNvsPCSNN}(A).
It's important to construct a feedback structure in order to backpropagate the intermediate gradient signal from $\delta_{k}^{l+1}$ to $\delta_{j}^{l}$. This means that the weights of the feedback network must correspond with those of the feed-forward network.

\section{Method}
In this section, we present our novel learning framework for SNNs, which incorporates predictive coding and local Hebbian synaptic plasticity mechanisms for both classification and regression tasks.

\subsection{PC-SNN for Classification}

\begin{figure*}[ht]f
  \centering
  \includegraphics[width=\linewidth]{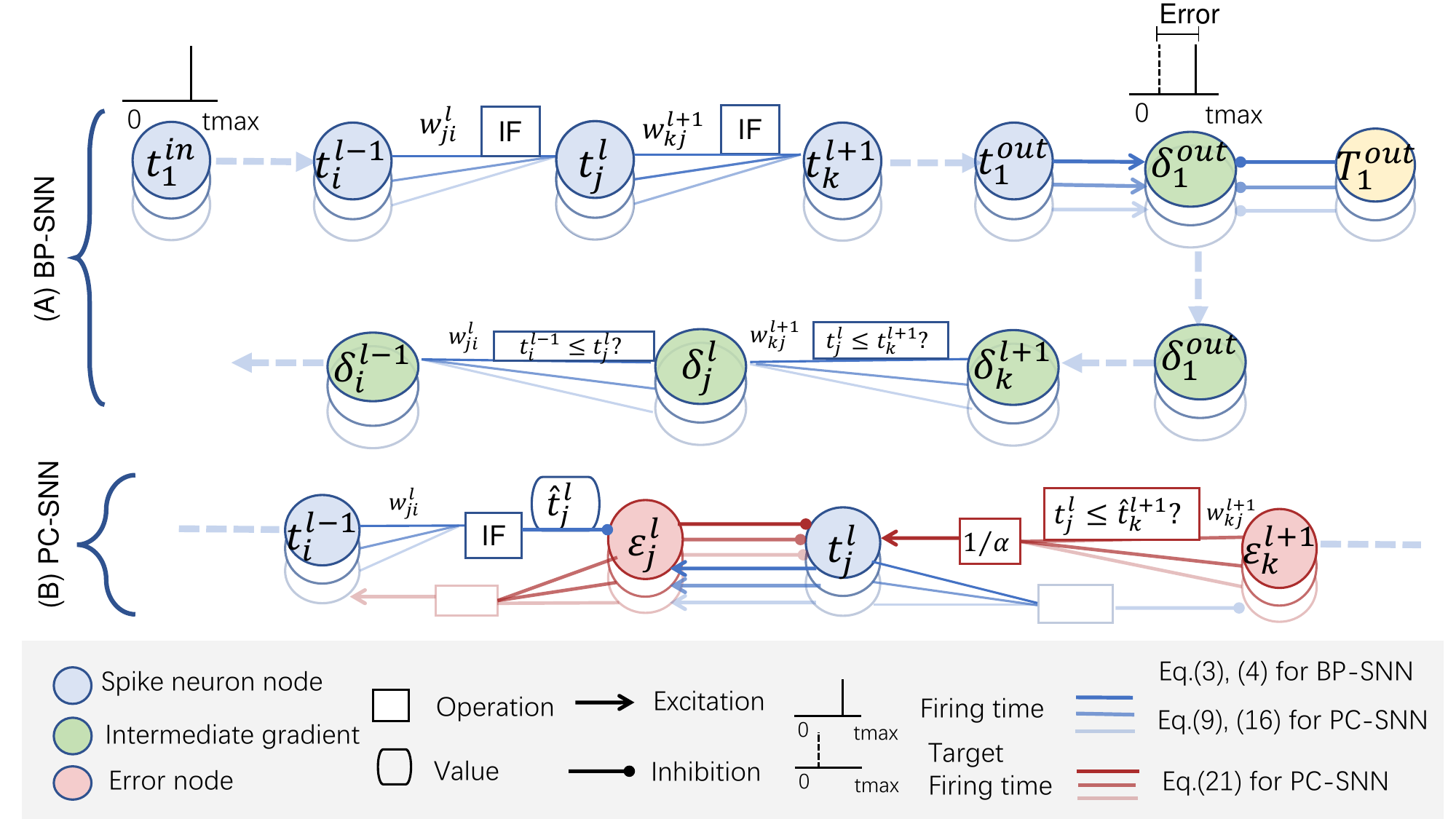}
  \caption{(A): SNN trained with BP (BP-SNN).
  % This is a two-layer spiking neural network. Input layer encodes the pixel values into spike trains by temporal encoding. IF neurons in the hidden and output layer transmit information by processing the received spike train and responding to it. Backpropagation compares the actual firing time obtained by the output layer with the target firing time, and propagates the gradient backward layer by layer. As we can see, constructing a feedback structure is an essential step to backpropagate the intermediate gradient signal from $\delta_{k}^{l+1}$ to $\delta_{j}^{l}$, indicating that each weight of the feedback network must be brought into correspondence with
  % the feed-forward network. 
  {It shows a two-layer SNN, and the input layer encodes pixel intensity values into spike trains using a temporal encoding scheme. IF neurons within the hidden and output layers process these received spike trains to transmit information. The network learns via a backpropagation algorithm that compares the actual output firing times against target firing times and propagates the resulting error gradient backward through the network's layers. The effective backpropagation of the intermediate gradient signal from $\delta_{k}^{l+1}$ to $\delta_{j}^{l}$ necessitates the construction of a feedback structure. This implies a critical constraint: each weight in the feedback network must correspond symmetrically to its counterpart in the feed-forward network.}
  (B): Predictive coding structure of SNN (PC-SNN). 
  % Both Predictions and prediction errors are updated in parallel mode using only local information.
  {The update mechanism for both predictions and prediction errors operates in a parallel manner, exclusively utilizing local information.}
   Note: Excitation denotes additive operator and Inhibition denotes subtraction operator.}
  \label{fig:SNNvsPCSNN}
\end{figure*}

\subsubsection{Encoding strategy}
During SNN training, the pixel intensities of images can be converted to spikes by multiple strategies. We use a Time-To-First-Spike time-based (TTFS) coding strategy \citep{rathi2020ttfs} that encodes the firing time of neurons, rather than the analog output of an ANN neuron through spiking rates. 

% Specifically, each neuron can only be activated once within a predefined time interval, and those neurons that fire first are considered the most active. In short, the larger the pixel value, the earlier the spike fires.
{Within a predefined temporal interval, each neuron is restricted to a single activation event. The timing of this activation is critical, as neurons that fire earliest are interpreted as having the highest activity levels. Consequently, an inverse relationship exists between the magnitude of a pixel value and the latency of the corresponding spike, with higher values precipitating earlier spike emissions.}

\subsubsection{Neuron Dynamics}
We integrate the potentials of neurons in the hidden and output layers by accumulating received spikes from presynaptic neurons through dendritic connections without leakage. Each presynaptic spike corresponds to a synaptic weight that raises the neuron's potential during integration, leading to the emission of signals when the internal potential reaches a predefined threshold. If a neuron's potential does not reach the threshold, it remains inactive. During training, it is essential to determine the firing times of all neurons. 
Therefore, {Consequently, if a neuron fails to reach its firing threshold during the training phase, its activation time is formally assigned the maximum value of the temporal window. Furthermore, this study adopts the dynamic target firing time methodology presented in S4NN \citep{kheradpisheh2020S4NN}, which is a technique that incorporates the actual neuronal activation times.}
% if a neuron does not reach the threshold during training, its activation time is artificially set to the maximum time. And we followed the dynamic target firing time setting in S4NN \citep{kheradpisheh2020S4NN} which incorporates the actual activation time.

\subsubsection{PC-SNN framework}
To address the non-differentiable nature of SNNs and achieve local synaptic plasticity learning, we adpot predictive coding framework with inference and learning schemes inspired by the Expectation-Maximization (EM) algorithm \citep{dempster1977maximum}. 

% The E-step involves finding the conditional expectation of the causes (inference), while the M-step determines the maximum likelihood value of the parameters (learning). 
{The Expectation (E) step, which constitutes the inference stage, is responsible for computing the conditional expectation of the latent variables. In contrast, the Maximization (M) step represents the learning stage, wherein the model's parameters are updated to their maximum likelihood estimates.}
% Fig.~\ref{fig:SNNvsPCSNN}(B) shows a demonstration of an SNN predictive coding model in parallel with the architecture of BP-SNN shown in Fig.~\ref{fig:SNNvsPCSNN}(A). 
As shown in Fig.~\ref{fig:SNNvsPCSNN}(B), an SNN predictive coding model is presented alongside the BP-SNN architecture illustrated in Fig.~\ref{fig:SNNvsPCSNN}(A).

\textbf{Probabilistic Model: }
In the probabilistic point of view, we consider the firing time of IF neurons as a random variable. Let $\textbf{t}^{l-1}$ be a vector of firing times on layer $l-1$, and $t_{j}^{l}$ be the firing time of neuron $j$ in layer $l$. Following the design of hierarchical models used in predictive coding \citep{friston2003learning}, we assume that adjacent layers' random variables satisfy the following relationships:
\begin{equation}\label{normal}
  p\left(t_{j}^{l} \mid \textbf{t}^{l-1}\right)=\mathcal{N}\left(t_{j}^{l} ;\hat{t}_{j}^{l}, \Sigma_{j}^{l}\right)
\end{equation}
here we adopt the a standard predictive coding probability density function notation $\mathcal{N}(x; \mu, \sigma^2)$ of a normal distribution evaluated at $x$, with mean $\mu$ and variance $\sigma^2$, and the mean probability density value $\hat{t}_{j}^{l}$ is a non-linear function of lower-layer IF neuron activity and is determined as the moment when membrane potential $V_{j}^{l}(t)$ first crosses the threshold, with variance $\Sigma_{j}^{l}$ remaining constant:
\begin{subequations}
\begin{flalign}
&  V_{j}^{l}(t)=\sum_{i} w_{j i}^{l} \sum_{\tau=1}^{t} \delta\left(\tau-t_{i}^{l-1}\right)\label{voltage} \\
 & \hat{t}_{j}^{l}=\min\left\{t: V_{j}^{l}(t)>=\vartheta\right\} \label{firing_time}
\end{flalign}
\end{subequations}
with a constant threshold $\vartheta$ and it is equal for all neurons in this work.

\textbf{Inference: }
The E-step, also known as Inference, boosts the probability $P$ based on the anticipated cause to achieve a reliable approximation of the recognition distribution indicated by network parameters. In this scenario, we use inference to identify the most probable random variable for neuron activity. To accomplish this, we must maximize the probability function given our inputs (for technical specifics, refer to \citep{friston2005theory}):
\begin{equation}\label{F1}
  F=P\left(\textbf{t}^{1}, \ldots, \textbf{t}^{l_{\max }} \mid \textbf{t}^{0}\right)
\end{equation}
%  to determine the activity of each IF neuron and converge $F$ by iteratively modifying $t_{i}^{l}$ values. To simplify the calculation, we express the probability function $F$ in the form of the logarithm function (because of the monotonicity of the logarithm function, maximizing the logarithm function has the identical impact of  maximizing the $F$ function itself):
{to determine the activity of each IF neuron and to achieve convergence for the objective function $F$by iteratively modifies the neuronal firing times, denoted as $t_{i}^{l}$. For reasons of computational simplicity, the optimization is performed on the logarithmic representation of this function. This transformation is justified by the monotonic property of the logarithm, which ensures that maximizing the log-function is equivalent to maximizing the function $F$ itself.}

\begin{equation}\label{F2}
  F=\ln\left(P\left(\textbf{t}^{1}, \ldots, \textbf{t}^{l_{\max }} \mid \textbf{t}^{0}\right)\right).
\end{equation}
Since we assumed that the random variables on one layer depend only on that of the previous layer (first-order Markov property), we can rewrite the objective function as:

\begin{equation}\label{F3}
  F=\sum_{l=1}^{l_{\max }} \ln \left(P\left(\textbf{t}^{l} \mid \textbf{t}^{l-1}\right)\right).
\end{equation}

% Because the activity of each neuron in one layer is independent of each other, we can write the probability of the random variable vector as a product of single random variable probabilities. Substituting \eqref{normal} and the expression for a normal distribution into the above equation \eqref{F3}, we obtain:
{Given the mutual independence of neuronal activities within a single layer, the joint probability of the random variable vector can be expressed as the product of the probabilities of the individual random variables. Consequently, by substituting equation \eqref{normal} and the probability density function for a normal distribution into equation \eqref{F3}, the following expression is derived:}
\begin{equation}\label{F4}
  F=\sum_{l=1}^{l_{\max }} \sum_{j=1}^{n^{l}}\left[\ln \left(\frac{1}{\sqrt{2 \pi} \Sigma_{j}^{l}}\right)-\frac{\left(t_{j}^{l}-\hat{t}_{j}^{l}\right)^{2}}{2 \Sigma_{j}^{l}}\right]
\end{equation}
where $n^l$ represents the number of neurons in the $l^{th}$ layer.
% Then, ignoring constant terms related to the variance, we can organize the objective function as
{By omitting constant terms associated with the variance, the objective function can be reformulated as follows:}
\begin{equation}\label{a}
  F=-\frac{1}{2} \sum_{l=1}^{l_{\max }} \sum_{j=1}^{n^{l}} \frac{\left(t_{j}^{l}-\hat{t}_{j}^{l}\right)^{2}}{\Sigma_{j}^{l}}.
\end{equation}

To optimize the objective function mentioned above by determining the values of $t_{j}^{l}$, we can modify them proportionally to the gradient. When calculating the derivative of $F$ with respect to $t_{j}^{l}$, we notice that each value affects $F$ in two ways: it appears explicitly in \eqref{a}, and it also impacts $\hat{t}_{k}^{l+1}$ according to \eqref{voltage}. Therefore, the derivative consists of two terms:
\begin{equation}\label{equ:F_deri}
  \frac{\partial F}{\partial t_{j}^{l}}=-\frac{t_{j}^{l}-\hat{t}_{j}^{l}}{\Sigma_{j}^{l}}+\sum_{k=1}^{n^{l+1}} \frac{t_{k}^{l+1}-\hat{t}_{k}^{l+1}}{\Sigma_{k}^{l+1}} \cdot \frac{\partial \hat{t}_{k}^{l+1}}{\partial t_{j}^{l}}.
\end{equation}
Let's denote
%$\varepsilon_{j}^{l}=({t_{j}^{l}-\hat{t}_{j}^{l}})/{\Sigma_{j}^{l}}$
\begin{equation}\label{error}
  \varepsilon_{j}^{l}=({t_{j}^{l}-\hat{t}_{j}^{l}})/{\Sigma_{j}^{l}}
\end{equation}
and this error node computes the difference between the current value of $t_{j}^{l}$ and the mean of $\hat t_{j}^{l}$ predicted by the lower layer. Then, from \eqref{equ:F_deri}, we have:
\begin{equation}\label{variational}
   \frac{\partial F}{\partial t_{j}^{l}}= - \varepsilon_{j}^{l} + \sum_{k=1}^{n^{l+1}} \varepsilon_{k}^{l+1} \cdot \frac{\partial \hat{t}_{k}^{l+1}}{\partial t_{j}^{l}}.
\end{equation}
To compute the derivative $\frac{\partial \hat{t}_{k}^{l+1}}{\partial t_{j}^{l}}$, we unfold this term according to chain rule:
\begin{equation}\label{b}
  \frac{\partial \hat{t}_{k}^{l+1}}{\partial t_{j}^{l}}=\frac{\partial \hat{t}_{k}^{l+1}}{\partial V_{k}^{l+1}(t)} \cdot \frac{\partial V_{k}^{l+1}(t)}{\partial t_{j}^{l}}.
\end{equation}

Assuming a small enough region around $t=\hat{t}_{k}^{l+1}$, we approximate the function $V_{k}^{l+1}(t)$ as a linear function of $t$ for the first factor in \eqref{b} \citep{spikeprop}. We denote the local derivative of $V_{k}^{l+1}(t)$ with respect to $t$ as $\alpha$, which is assumed to be a fixed positive constant in this paper. Since the threshold can only be reached on the rising edge of the membrane potential for the first time, it follows that $V_{k}^{l+1}(t)$ increases over time around $t=\hat{t}_{k}^{l+1}$. If there is an increment in  $V_{k}^{l+1}(t)$ around  $ t= \hat { t } _ { k } ^ { l + 1 }$, then intuitively, it will reach its threshold earlier and cause a decrease in firing time $\hat{t}_{k}^{l+1}$ . Therefore, $\frac{\partial \hat{t}_{k}^{l+1}}{\partial V_{k}^{l+1}( t )}<0$. To approximate $\frac{\partial \hat{ t } _ { k } ^ { l + 1 }} {\partial V_ { k } ^ { l + 1 }( t )}$, considering the derivative of inverse function ${V_{{\rm{k}}}^{{\rm{l}} + 1}}( {{\rm{t}}} )$, we have:
\begin{equation}\label{1}
  \frac{\partial \hat{t}_{k}^{l+1}}{\partial V_{k}^{l+1}(t)}=-\frac{1}{\frac{\partial V_{k}^{l+1}(t)}{\partial t}} = -\frac{1}{\alpha}.
\end{equation}

For the second factor in \eqref{b}, it can be simplified by considering equation \eqref{voltage}. When $t_{j}^{l}$ is reduced, $V_{k}^{l+1}(t)$ increases earlier in time by $w_{kj}^{l+1}$. Therefore, we can approximate $\frac{\partial V_{k}^{l+1}(t)}{\partial t_{j}^{l}}$ as $-w_{kj}^{l+1}$ only if $\left[t_{j}^{l}\leq\hat{t}_{k}^{l+1}\right]$. By substituting these derivative terms into equation \eqref{b}, we obtain:
\begin{equation}\label{}
   \frac{\partial \hat{t}_{k}^{l+1}}{\partial t_{j}^{l}}=\begin{cases} \frac{1}{\alpha} \cdot w_{kj}^{l+1} & \text { if } t_{j}^{l} \leq \hat{t}_{k}^{l+1} \\ 0 & \text { otherwise }\end{cases}
\end{equation}

Finally, we have derived the following rule that describes how $t_{j}^{l}$ changes over time:
\begin{equation}\label{eq:firingtimeupdate}
   \frac{\partial F}{\partial t_{j}^{l}}= - \varepsilon_{j}^{l} + \sum_{k=1}^{n^{l+1}} \varepsilon_{k}^{l+1} \cdot \frac{1}{\alpha} \cdot w_{kj}^{l+1} \left[t_{j}^{l} \leq \hat{t}_{k}^{l+1}\right]
\end{equation}
and the temporal activity $t_{j}^{l}$ is updated depending only on local nodes and weights.

\textbf{Learning parameters:}
During the training-time inference step, the firing times of neurons in the final layer are set to correspond to the target output firing times ($t_{i}^{o}=T_{i}^{o}$). Subsequently, the firing times for all neurons in the antecedent layers, where $l \in\left\{1, \ldots, l_{\max }-1\right\}$, are updated in accordance with the previously established method detailed in Equation \eqref{eq:firingtimeupdate}.
From a neurobiological perspective, both learning and inference aim to minimize free energy $F$ in exactly the same way according to Friston's theory \citep{friston2005theory}. As we modify $w_{kj}^{l+1}$ proportionally to its objective function gradient, our network gradually approaches a steady state. Eventually, updating all synaptic weights will lead to predicting desired outputs. It should be noted that modifying $w_{kj}^{l+1}$ affects $\hat{t}_{k}^{l+1}$ and thus influences function value $F$ formulated in \eqref{a}:
\begin{align}\label{eq:weightupdate}
&\frac{\partial F}{\partial w_{kj}^{l+1}}=\frac{t_{k}^{l+1}-\hat{t}_{k}^{l+1}}{\Sigma_{k}^{l+1}} \cdot \frac{\partial \hat{t}_{k}^{l+1}}{\partial w_{kj}^{l+1}} \nonumber \\
&=\varepsilon_{k}^{l+1} \cdot \frac{\partial \hat{t}_{k}^{l+1}}{\partial V_{k}^{l+1}(t)} \cdot \frac{\partial V_{k}^{l+1}(t)}{w_{kj}^{l+1}} \nonumber \\
&=\varepsilon_{k}^{l+1} \cdot-\frac{1}{\alpha} \cdot \sum_{\tau=1}^{\hat{t}_{k}^{l+1}} S_{j}^{l}(\tau)
\end{align}
where $\sum_{\tau=1}^{\hat{t}_{k}^{l+1}}S_{j}^{l}(\tau)=1$ if $t_{j}^{l} \leq \hat{t}_{k}^{l+1} $ else 0.

% According to \eqref{eq:firingtimeupdate} and \eqref{eq:weightupdate}, the change in neuron activity $t_{j}^{l}$ and synaptic weight $w_{kj}^{l+1}$ between layer $l$ and $l+1$ are both proportional to the product of temporal quantities encoded on these two adjacent layers, which can be directly obtained by local connection. The update of its weight ($w_{kj}^{1+1}$) depends only on temporal activities of presynaptic and postsynaptic nodes, i.e., $\varepsilon_{k}^{l+1}$ and $S_{j}^{l}(t)$, satisfying biologically-plausible local plasticity. This is in contrast to the expression of backpropagation, where the  weight changes depending on intermediate variables that are passed from the back to the front through a complex function of activities and weights. We refer to the changes in \eqref{eq:weightupdate} as local Hebbian synaptic plasticity in a manner that the weight change is a simple product of  temporal activities of presynaptic and postsynaptic spiking neurons. We use pseudocode to describe the learning and prediction process in Algorithm~\ref{algorithm1} and Algorithm~\ref{algorithm2}, respectively.

Based on Equations \eqref{eq:firingtimeupdate} and \eqref{eq:weightupdate}, adjustments to both the neuronal activity $t_{j}^{l}$ and the synaptic weight $w_{kj}^{l+1}$ between two adjacent layers, $l$ and $l+1$, are proportional to the product of temporal quantities encoded within these layers. This calculation relies solely on information available through local connections. The weight update, $w_{kj}^{l+1}$, is governed exclusively by the temporal activities of the presynaptic ($S_{j}^{l}(t)$) and postsynaptic ($\varepsilon_{k}^{l+1}$) neurons, a mechanism that adheres to the principles of biologically plausible local plasticity. Similar to most supervised learning frameworks, our algorithm requires a complete forward pass to determine spike times before the learning phase, after which target firing times at the output layer guide the inference process. However, the inference and learning computations themselves are fundamentally local, where each neuron's update depends only on its own error node and signals from immediately adjacent neurons. Importantly, since these local computations do not require information from distant layers, the updates across all hidden layers can be executed simultaneously in parallel, as explicitly indicated in Algorithm~\ref{algorithm1}. This stands in contrast to backpropagation, where gradient computation requires sequential propagation through all layers via the chain rule.
This local dependency is in direct contrast to the backpropagation algorithm, where weight modifications are contingent upon intermediate variables propagated backward through the network via complex functions of activities and weights. The update rule in Equation \eqref{eq:weightupdate} is therefore defined as local Hebbian synaptic plasticity, given that the change in synaptic strength is a direct product of the temporal activities of interconnected presynaptic and postsynaptic spiking neurons. The learning and prediction processes are formally delineated using pseudocode in Algorithm~\ref{algorithm1} and Algorithm~\ref{algorithm2}, respectively.

% \subsubsection{PC-SNN analysis}
% We assume that the dataset $\mathcal{Z}=(\mathcal{X}, \mathcal{Y})$ is complete, but only $\mathcal{X}$ is observed. In this context, we consider an input spike $S^0(t)$ and a target firing time $T^{o}$ as $\mathcal{X}$, which is given. The latent variables $\textbf{t}^l$ on layer $l \in\left\{1, \ldots, l_{\max }-1\right\}$ are considered as $\mathcal{Y}$. We denote the complete-data log likelihood by $l(\theta ; \mathcal{X}, \mathcal{Y})$, where $\theta$ represents the unknown parameter vector for which we want to find the Maximum Likelihood Expectation. Here, $l(\theta ; \mathcal{X}, \mathcal{Y})$ corresponds to our objective function $F$, following a nonlinear model under Gaussian assumptions (Equation \eqref{normal}). The parameter set of the spiking neural network is denoted by $\theta$. According to Bishop's EM algorithm \citep{bishop2006pattern}, one has following steps:

{The analytical framework presupposes a complete dataset denoted by $\mathcal{Z}=(\mathcal{X}, \mathcal{Y})$, of which only the subset $\mathcal{X}$ is directly observed. Within this framework, the observed data $\mathcal{X}$ comprises an input spike train $S^0(t)$ and a corresponding target firing time $T^{o}$. The latent variables, constituting the unobserved data $\mathcal{Y}$, are the firing times $\textbf{t}^l$ for each hidden layer $l \in\left\{1, \ldots, l_{\max }-1\right\}$. The complete-data log-likelihood is represented by $l(\theta ; \mathcal{X}, \mathcal{Y})$, where $\theta$ is the unknown parameter vector for which the Maximum Likelihood Estimate is sought. This log-likelihood function corresponds to the objective function $F$, which is formulated based on a nonlinear model with Gaussian assumptions as defined in Equation \eqref{normal}. Following the standard Expectation-Maximization (EM) algorithm as detailed by \cite{bishop2006pattern}, the process involves the subsequent iterative steps.}

\textbf{E-Step}: 
% The E-step of the EM algorithm computes the expected value of $l(\theta ; \mathcal{X}, \mathcal{Y})$ given the observed data $\mathcal{X}$, and the current parameter estimate $\theta_{\text {old }}$. In particular, we define:
{The E step evaluates the conditional expectation of the log-likelihood function $l(\theta ; \mathcal{X}, \mathcal{Y})$. This calculation is performed based on the observed data $\mathcal{X}$ and the current parameterization $\theta_{\text{old}}$. Formally, this is expressed as:}
\begin{equation}\label{E}
  \begin{aligned}
F\left(\theta ; \theta_{\text {old }}\right) &:=\mathrm{E}\left[l(\theta ; \mathcal{X}, \mathcal{Y}) \mid \mathcal{X}, \theta_{\text {old }}\right] \\
&=\int l(\theta ; \mathcal{X}, y) p\left(y \mid \mathcal{X}, \theta_{o l d}\right) d y
\end{aligned}
\end{equation}
where  $p\left(\mathcal{Y} \mid \mathcal{X}, \theta_{o l d}\right)$ is the conditional density of $\mathcal{Y}$ given the observed data, $\mathcal{X}$. The goal of the E-step is to determine the distribution of $p\left(\mathcal{Y} \mid \mathcal{X}, \theta_{o l d}\right)$, which can also be defined as Gaussian. In the present context of the hierarchical Gaussian generative model, we define $p\left(\mathcal{Y} \mid \mathcal{X}, \theta_{o l d}\right) = \prod_1^{l_{max}-1} \mathcal{N}\left(\mathcal{Y}^l; \mu^{l}, \sigma^{l}\right), l=1:l_{max}-1$. This intractable posterior can be approximated with variational inference as proved in \citep{millidge2020predictive}. The final updating form of neural activities in \citep{millidge2020predictive} coincides with maximizing  $F\left(\theta ; \theta_{\text {old }}\right)$ with respect to $\text{t}^l$ in the inference process, as shown in \eqref{variational}.

\textbf{M-Step}: The M-step consists of maximizing over $\theta$ the expectation computed in \eqref{E}, that is, we set:
\begin{equation}\label{M}
 \theta_{\text {new }}:=\max _\theta F\left(\theta ; \theta_{\text {old }}\right).
\end{equation}
Then, we set $\theta_{old} = \theta_{new}$ for iteration updating. This solution of M-step is shown as the parameters updated in \eqref{eq:weightupdate}. The two steps are repeated as necessary until the sequence of $\theta_{new}$’s converges.

In practice, as shown in Algorithm 1, we implement a simultaneous update scheme where both E-step (inference of neural activities) and M-step (learning of weights) are performed iteratively within the same convergence loop. This approach is computationally efficient while maintaining the theoretical guarantees of EM optimization, as both updates maximize the same objective function F.

\begin{algorithm}[hbt!]
\caption{Learning with predictive coding}\label{algorithm1}
\For{all training Data}
{
Temporal coding: $\textbf{S}^{0}\leftarrow \textbf{S}^{input}$\;

\tcp{Forward pass to obtain actual output firing times}
    \For{$l = 1$ \KwTo $l_{\max}$}{
        $V_{j}^{l}(t)=\sum_{i} w_{j i}^{l} \sum_{\tau=1}^{t} S_{i}^{l-1}(\tau)$ (Eq.3)\;
        ${t}_{j}^{l}=\min\left\{t: V_{j}^{l}(t)>=\vartheta\right\}$ (Eq.4)\;
        % Compute $V_j^l(t)$ using Eq. (3)\;
        % Compute $t_j^l$ using Eq. (4)\;
    }
    
    \tcp{Set dynamic target firing time}
    $\tau \leftarrow \min\{t_j^o \mid 1 \leq j \leq C\}$\;
    Set $T^o$ according to Eq. (5)\;
    % $t^{l_{\max}} \leftarrow T^o$\;
    Setting dynamic target firing time: $\textbf{t}^{l_{max}}\leftarrow \textbf{T}^{o}$\;

\While{not convergence}
{
  \tcp{E-step(Inference)}
  \tcp{Can be executed in parallel}
    \For{$l = 1$ \KwTo $l_{\max}-1$}{  
      \For{each neuron $i$ in layer $l$}{
        $\varepsilon_{i}^{l}=\frac{t_{i}^{l}-\hat{t}_{i}^{l}}{\Sigma_{i}^{l}}$ (Eq.17) \;
        $\dot{t}_{i}^{l}=-\varepsilon_{i}^{l} + \sum_{j=1}^{n^{l+1}}\varepsilon_{j}^{l+1}\cdot\frac{1}{\alpha}\cdot w_{ji}^{l+1}\left[t_{i}^{l} \leq \hat{t}_{j}^{l+1}\right]$ (Eq.22) \;
        $t_{i}^{l}\leftarrow t_{i}^{l}+ \dot{t}_{i}^{l}$\;
      }
    }
  \tcp{M-step(Update weights)}
  \tcp{Can be executed in parallel}
    \For{$l = 0$ \KwTo $l_{\max}-1$}{
        \For{each synapse $w_{ji}^{l+1}$ between layer $l$ and $l{+}1$}{
          $\dot{w}_{j i}^{l+1}=\varepsilon_{j}^{l+1} \cdot-\frac{1}{\alpha} \cdot \sum_{\tau=1}^{\hat{t}_{j}^{l+1}} S_{i}^{l}(\tau)$ (Eq.23)\;
          $w_{j i}^{l+1}\leftarrow w_{j i}^{l+1}+\eta \dot{w}_{j i}^{l+1}$\;
        }
    }
}
}
\end{algorithm}

\begin{algorithm}[hbt!]
\caption{Prediction after training}\label{algorithm2}
\For{all testing Data}
{
Temporal coding: $\textbf{S}^{0}\leftarrow \textbf{S}^{input}$\;
\tcp{Forward pass to compute predictions}
\For{$l = 1$ \KwTo $l_{\max}$}{
        $V_{j}^{l}(t)=\sum_{i} w_{j i}^{l} \sum_{\tau=1}^{t} S_{i}^{l-1}(\tau)$ (Eq.3)\;
        ${t}_{j}^{l}=\min\left\{t: V_{j}^{l}(t)>=\vartheta\right\}$ (Eq.4)\;
}
% $V_{j}^{l}(t)=\sum_{i} w_{j i}^{l} \sum_{\tau=1}^{t} S_{i}^{l-1}(\tau)$\;
% ${t}_{j}^{l}=\min\left\{t: V_{j}^{l}(t)>=\vartheta\right\}$\;
}
\end{algorithm}

\subsection{ASRNN}

\subsubsection{ASRNN for regression}

\cite{yin2021accurate} demonstrate that state-of-the-art performance for sequential and temporal tasks in SNNs is attainable through the use of an Adaptive Spiking Recurrent Neural Network, which is capable of learning temporal dynamics.
This is achieved using a new substitute differential, the Gaussian distribution, in backpropagation. The results are comparable to conventional RNNs, with a theoretical energy advantage of 1 to 3 orders of magnitude.
This advantage grows with task complexity, requiring larger networks for precise solutions. Theoretically, the ASRNN model supports long-term and short-term memory, analogous to LSTM and other traditional neural network memory units.
In this paper, the ASRNN is extended to continuous-time regression prediction for the first time, improving upon the previous SRM-based model.
Originating from the LIF spiking neuron, Adaptive SRNN integrates input current $I(t)$ in a leaky manner and fires an action potential when its membrane potential $\mu(t)$ crosses a dynamic threshold $\theta$. The threshold increases after each spike and decays exponentially with time constant $\tau_{adp}$. The differential equation is expressed as follows:

\begin{eqnarray}
  && \tau_{m} \frac{d \mu}{d t}=-\mu(t)+R_{m} I(t)-\tau_{m} \theta(t) S(t) \label{b_2} \\
   && \tau_{adp}\frac{d \theta}{d t}=-\left(\theta(t)-b_{0}\right)+\beta S(t) \label{c}
\end{eqnarray}

From a dynamical perspective, the adaptive threshold in ASRNN plays a stabilizing, homeostatic role rather than simply adding extra complexity to the neuron model. Each spike transiently increases the threshold and thereby reduces the probability of subsequent firings, while the exponential decay with time constant $\tau_{adp}$ gradually relaxes the threshold back to its baseline $b_0$ when the neuron remains silent. This spike-triggered negative feedback prevents runaway firing and limits large excursions of the membrane potential, which in turn keeps the recurrent dynamics in a bounded operating regime. As a result, ASRNN neurons tend to exhibit sparse and temporally structured activity instead of dense, highly synchronous spikes.

In simulation, the continuous neuron model is discretized, we set the time interval as $dt=1ms$, combined with the first-order Taylor expansion, The $n_{l}$ layer feed-forward spiking convolutional neural network based on ASRNN is modeled as follows:
\begin{subequations}
  \begin{flalign}
  & \boldsymbol{s}^{0}(t) =\boldsymbol{s}_{\text {in }}(t) \\
  & \boldsymbol{\eta}^{l+1}(t)=\rho \boldsymbol{\eta}^{l+1}(t-1)+(1-\rho) \boldsymbol{s}^{l+1}(t-1)\\
  & \boldsymbol{\vartheta}^{l+1}(t) =b_{0}+\beta \boldsymbol{\eta}^{l+1}(t)\\
  & \boldsymbol{I}^{l+1}(t) = \boldsymbol{W}^{l+1}\boldsymbol{s}^{l}(t) \\
  &\boldsymbol{u}^{l+1}(t)=\alpha \boldsymbol{u}^{l+1}(t-1)+(1-\alpha) R_{m}\boldsymbol{I}^{l+1}(t)\label{membrane}\\
  &\quad\quad\quad\quad -\boldsymbol{\vartheta}^{l+1}(t) \boldsymbol{s}^{l+1}(t-1)  \nonumber \\
  & \boldsymbol{s}^{l}(t) =\sum \delta\left(t-t^{f}\right)\\
  & t^{f}  \in\left\{t \mid \boldsymbol{u}^{l}(t)=\vartheta\right\} \\
  & \boldsymbol{\omega}(t) =  \boldsymbol{u}^{o}(t)
  \end{flalign}
\end{subequations}
where $\alpha=\exp \left(-d t / \tau_{m}\right)$ is the single-timestep decay of the membrane potential with time-constant $\tau_{m}$, $\rho=\exp \left(-d t / \tau_{a d p}\right)$ is the single-timestep decay of the threshold with time-constant $\tau_{adp}$ , $\boldsymbol{\vartheta}$ is a dynamical threshold comprised of a fixed minimal threshold $b_{0}$ and an adaptive contribution $\beta \boldsymbol{\eta}$; The parameter $\beta$ is a constant that controls the size of adaptation of the threshold.

\begin{figure*}[hbt!]
  \centering
  \subfigure[ASRNN-BPTT]{
    \includegraphics[width=0.86\textwidth]{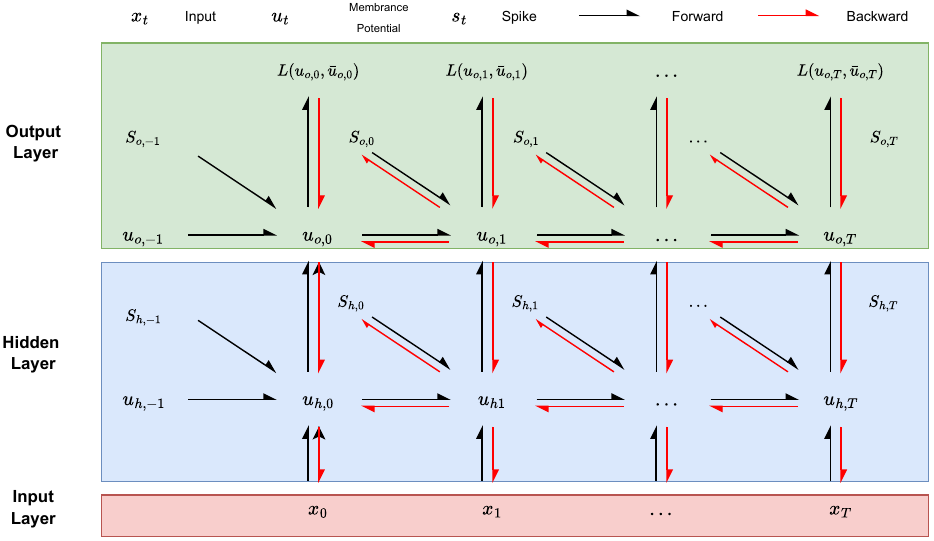}
    \label{fig:bp}
  }
  \subfigure[ASRNN-PCTT]{
    \includegraphics[width=0.8\textwidth]{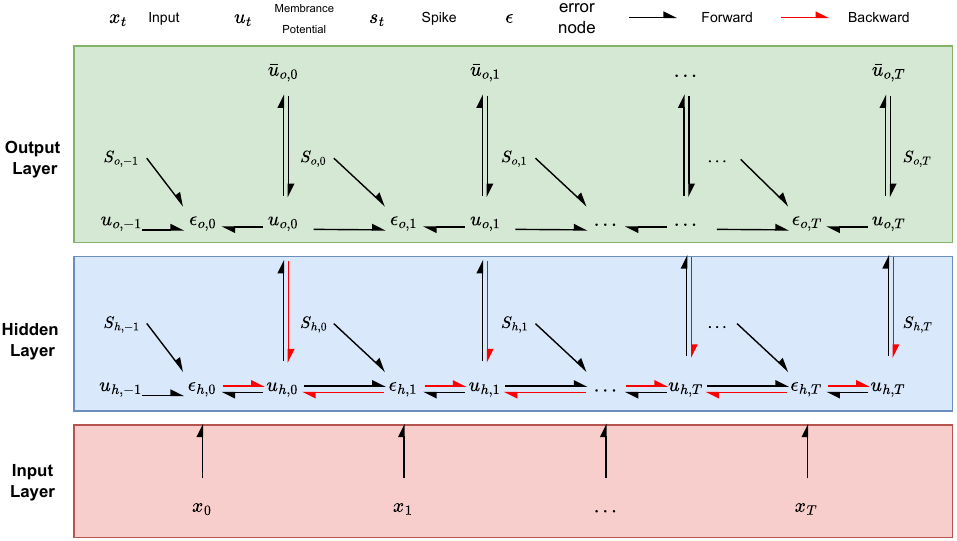}
    \label{fig:pc}
  }
  \caption{An diagram illustration between  ASRNN-BPTT and ASRNN-PCTT.}
  
\end{figure*}

\subsubsection{ASRNN-BPTT Algorithm}
In this section, we derive the ASRNN-BPTT algorithm and analyze the derivation results. By BPTT, the difference between the prediction and target is transmitted from the output layer back to the input layer, including the input layer at the past time, optimizing weights and parameters by gradient descent. Conceptually, BPTT expands the network at all input time steps.

The discontinuous nature in spiking neurons makes it difficult to apply the chain rule to calculate the back propagation gradient. In practice, replacing discontinuous gradients with surrogate gradients, has been shown to be effective, making it possible to implement spiking neural networks training on mainstream deep learning frameworks such as PyTorch and Tensorflow. A variety of alternative gradient functions were proposed and evaluated, including multi-Gaussian function, Gaussian function, linear function and SLayer function.For these functions, however, the study showed no significant difference in performance. We use Gaussian function \citep{yin2020effective} in this paper: $\hat{f} _{s}^{\prime}\left(u_{t}\right)=\mathcal{N}\left(u_{t} \mid \theta, \sigma^{2}\right)$.

In the event camera angular velocity prediction task, we need to generate an output at each step $t$.The loss function $L$ is defined as the Euclidean distance of the predicted angular velocity $\boldsymbol{\omega}(t)$ and the ground truth angular velocity $\bar{\boldsymbol{\omega}}(t)$ over time:
\begin{equation}\label{loss}
  L=\frac{1}{T_{1}-T_{0}} \int_{T_{0}}^{T_{1}} \sqrt{\boldsymbol{e}(t)^{\top} \boldsymbol{e}(t)} \mathrm{d} t
\end{equation}
where $\boldsymbol{e}(t)=\boldsymbol{\omega}(t)-\bar{\boldsymbol{\omega}}(t)$. The loss function estimates the prediction error of angular velocity in the whole simulation time, i.e. 100ms with
a time step of 1ms. So the discretized form of the loss function is obtained:
\begin{subequations}
 \begin{flalign}
  & L=\frac{1}{T_{1}-T_{0}} \sum_{t=T_{0}}^{T_{1}} L(t)\\
  & L(t)=\| \boldsymbol{u}^{o}(t)-\bar{\boldsymbol{u}}^{o}(t)\|
 \end{flalign}
\end{subequations}

By the chain rule, We get the derivative of the loss function with respect to the weight $w_{ji}^{l+1}$:
\begin{equation}\label{bptt}
  \frac{\partial L}{\partial w_{ji}^{l+1}}=\sum_{t=T_{0}}^{T_{1}} \frac{\partial L(t)}{\partial w_{ji}^{l+1}}=\sum_{t=T_{0}}^{T_{1}} \frac{\partial L(t)}{\partial \mu_{j}^{l+1}(t)} \cdot \frac{\partial \mu_{j}^{l+1}(t)}{\partial w_{j i}^{l+1}}
\end{equation}

For the first derivative term in the right side of \eqref{bptt}, we denote the error term as:
\begin{equation}
  \label{loss_2}
  \delta_{j}^{l+1}(t)=\frac{\partial L(t)}{\partial \mu_{j}^{l+1}(t)}=\sum_{k=1}^{n^{l+2}} \frac{\partial L(t)}{\partial \mu_{k}^{l+2}(t)} \cdot \frac{\partial \mu_{k}^{l+2}(t)}{\partial \mu_{j}^{l+1}(t)} \nonumber
\end{equation}
where $\frac{\partial \mu_{k}^{l+2}(t)}{\partial \mu_{j}^{l+1}(t)}= (1-\alpha) R_{m}w_{k j}^{l+2} \hat{f}_{s}^{\prime}\left(u_{j}^{l+1}(t)\right)$ according to spiking neuron model depicted in \eqref{membrane}. Therefore, the recursive calculation of the error term from the output layer to the current layer is obtained:
\begin{equation}\label{delta}
  \delta_{j}^{l+1}(t)=(1-\alpha) \hat{f}_{s}^{\prime}\left(u_{j}^{l+1}(t)\right) \sum_{k=1}^{n^{l+2}} \delta_{k}^{l+2}(t) w_{k j}^{l+2}
\end{equation}

For the second derivative term in the right side of \eqref{bptt}, it is necessary to go through all time steps before $t$ and calculate iteratively in time as follows:
\begin{equation}\label{mu_w}
  \frac{\partial \mu_{j}^{l+1}(t)}{\partial w_{j i}^{l+1}}=(1-\alpha) s_{i}^{l}(t)+ \frac{\partial u_{j}^{l+1}(t)}{\partial u_{j}^{l+1}(t-1)} \cdot \frac{\partial u_{j}^{l+1}(t-1)}{\partial w_{j i}^{l+1}} \nonumber
\end{equation}
Denote $\lambda_{j}^{l+1}(t)=\frac{\partial u_{j}^{l+1}(t)}{\partial u_{j}^{l+1}(t-1)}$, and the gradient backtracking of BPTT from the current time step $t$ to the previous time step can be obtained as follows:
\begin{equation}\label{mu_time}
  \frac{\partial \mu_{j}^{l+1}(t)}{\partial w_{j i}^{l+1}}=(1-\alpha) s_{i}^{l}(t)+ \lambda_{j}^{l+1}(t) \cdot \frac{\partial u_{j}^{l+1}(t-1)}{\partial w_{j i}^{l+1}}
\end{equation}

The calculation graph is shown in Fig. \ref{fig:bp}. As shown, the calculation of weight gradient requires backpropagating in both spatial and temporal dimensions. Specifically, the error term $\delta_{j}^{l+1}(t)$ needs to be propagated layer by layer from the output layer as \eqref{delta}, and the iterative gradient $\frac{\partial \mu_{j}^{l+1}(t)}{\partial w_{j i}^{l+1}}$ also needs to be calculated recursively from initial time step as \eqref{mu_time}. On the one hand, spatial backpropagation of the error term $\delta_{j}^{l+1}(t)$ uses too much global information, i.e. the activity of neurons that are not directly connected to the weight modified, instead of local information, thus does not satisfy the local plasticity in biology. On the other hand, backpropagation through time is a common phenomenon in recurrent neural networks, which may cause problems such as gradient explosion or gradient vanishing.

\subsubsection{ASRNN-PCTT Algorithm}

% In order to further improve the BPTT algorithm mentioned above, inspired by Predictive Coding theory in\citep{1999predictiveframework, friston2003learning, friston2005theory, whittington2017approximation}, we propose a new supervised learning algorithm, ASRNN-PCTT, and apply it to a dataset for predicting angular velocity in event cameras.

{Drawing upon the theoretical foundations of Predictive Coding \citep{1999predictiveframework, friston2003learning, friston2005theory, whittington2017approximation}, this study presents ASRNN-PCTT, a novel supervised learning algorithm developed to advance the aforementioned BPTT algorithm. The proposed model's effectiveness is validated on a dataset for the task of angular velocity prediction using event cameras.}

\textbf{Probabilistic Model: }
% From a probabilistic perspective, we model the membrane potential of neurons as a random variable and assume that the variables in adjacent layers satisfy the following relationships:
{The membrane potential of a neuron is conceptualized as a random variable. It is further postulated that the variables corresponding to adjacent layers are governed by the set of relationships delineated below:}
\begin{equation}\label{mu_normal}
  P\left(u_{i}^{l}(t) \mid \textbf{u}^{l-1}(t)\right)=\mathcal{N}\left(u_{i}^{l}(t) ;\hat{u}_{i}^{l}(t), \Sigma_{i}^{l}\right)
\end{equation}
where the mean value of probability distribution $\hat{u}_{i}^{l}(t)$ satisfies the following expression according to original Adaptive SRNN neuron model in \eqref{membrane}:
\begin{equation}\label{mu_hat}
  \hat{u}_{i}^{l}(t)=\alpha u_{i}^{l}(t-1)+(1-\alpha) R_{m} \sum_{h} w_{ih}^{l} s_{h}^{l-1}(t)-s_{i}^{l}(t-1) \theta_{i}^{l}(t)
\end{equation}

\textbf{Inference: }
% Given the spiking input from the event-camera, Inference process is to find the most likely neuron activity random variable. We need to maximize the probability function:
{For a given spiking input from the event camera, the objective of the inference process is to determine the maximum a posteriori estimate of the random variable representing neural activity. This estimation is accomplished by maximizing the following probability function:}
\begin{equation}
  F=\prod_{t=T_{0}}^{T_{1}} P\left(\boldsymbol{u}^{1}(t), \cdots, \boldsymbol{u}^{l_{\max }}(t) \mid \boldsymbol{u}^{0}(t)\right) \nonumber
\end{equation}

Converting objective function $F$ into logarithmic form, and the joint probability distribution is decomposed into products of probabilities, $F$ can be written as :
\begin{equation}\label{f}
  F=-\frac{1}{2} \sum_{t=T_{0}}^{T_{1}} \sum_{l=1}^{\operatorname{lmax}} \sum_{i=1}^{n^l} \frac{\left(u_{i}^{l}(t)-\hat{u}_{i}^{l}(t)\right)^{2}}{\Sigma_{i}^{l}}
\end{equation}

In calculating the derivative of the objective function \eqref{f} with respect to $u_{i}^{l}(t)$, it is observed that each random variable $u_{i}^{l}(t)$ affects the objective function in three ways. First, it appears explicitly in the formula \eqref{f}; Second, it implicitly influences objective function by affecting $\hat{u}_{j}^{l+1}(t)$; Third, it also affects the objective function by affecting $\hat{u}_{i}^{l}(t+1)$, as \eqref{mu_hat}. Therefore, the derivative of the objective function with respect to $u_{i}^{l}(t)$ contains the following three items:
\begin{equation}
    \frac{\partial F}{\partial u_{i}^{l}(t)}=-\varepsilon_{i}^{l}(t)+\sum_{j=1}^{n^{l+1}} \varepsilon_{j}^{l+1}(t) \cdot \frac{\partial \hat{u}_{j}^{l+1}(t)}{\partial u_{i}^{l}(t)}+\varepsilon_{i}^{l}(t+1) \frac{\partial \hat{u}_{i}^{l}(t+1)}{\partial u_{i}^{l}(t)}
\end{equation}
% where the local error node $\varepsilon_{i}^{l}(t)$ computes the difference between the current value of $u_{i}^{l}(t)$ and the mean value $\hat{u}_{i}^{l}(t)$ that predicted by the lower layers:
{where the local error node $\varepsilon_{i}^{l}(t)$ quantifies the discrepancy between the current value $u_{i}^{l}(t)$ and the mean value $\hat{u}_{i}^{l}(t)$, which is predicted based on the outputs from the preceding layers:}

\begin{equation}
    \varepsilon_{i}^{l}(t)=\frac{u_{i}^{l}(t)-\hat{u}_{i}^{l}(t)}{\Sigma_{i}^{l}} \nonumber
\end{equation}

In addition, denote $\lambda_{i}^{l}(t+1)=\frac{\partial \hat{u}_{i}^{l}(t+1)}{\partial u_{i}^{l}(t)}$ and infer that $\frac{\partial \hat{u}_{j}^{l+1}(t)}{\partial u_{i}^{l}(t)}=(1-\alpha)R_{m} w_{ji}^{l+1} \hat{f}_{s}^{\prime}\left(u_{i}^{l}(t)\right)$, the derivative of the objective function with respect to $u_{i}^{l}(t)$ can be summarized as follows:
\begin{equation}\label{F_mu}
\begin{split}
   \frac{\partial F}{\partial u_{i}^{l}(t)} =& -\varepsilon_{i}^{l}(t)+\sum_{j=1}^{n^{l+1}} \varepsilon_{j}^{l+1}(t) \cdot (1-\alpha)R_{m}w_{ji}^{l+1} \hat{f}_{s}^{\prime}\left(u_{i}^{l}(t)\right)\\
     & +\varepsilon_{i}^{l}(t+1) \lambda_{i}^{l}(t+1)
\end{split}
\end{equation}

\textbf{Learning parameters: }
During the training-time inference step, the random variable of the neurons on the output layer are set to the target angular velocity, i.e $\boldsymbol{u}^{o}(t)=\bar{\boldsymbol{u}}^{o}(t)$. The values of random variable of all neurons on layer $l$ ($l \in\left\{1, \ldots, l_{\max }-1\right\}$) and time $t$ ($t \in\left\{T_{0}, \ldots, T_{1}\right\}$) are adjusted in the way as \eqref{F_mu} for several steps until convergence. After that, the network parameter $w_{ji}^{l+1}$ is updated with information in steady state. Easy to see that the weight $w_{ji}^{l+1}$ affects the value of the objective function $F$ by influencing the mean of probability model, i.e. $\hat{u}_{j}^{l+1}(t)$, and we get:
\begin{equation}\label{F_w}
  \frac{\partial F}{\partial w_{ji}^{l+1}}=\sum_{t=T_{0}}^{T_{1}} \varepsilon_{j}^{l+1}(t) \cdot (1-\alpha)R_{m}s_{i}^{l}(t)
\end{equation}

The calculation graph of PCTT is shown in Fig.~\ref{fig:pc}. According to \eqref{F_mu} and \eqref{F_w}, it can be concluded that whether the update of membrane potential or the update of weight parameters, in terms of spatial network structure, only local error nodes and weight information directly adjacent to the modified quantity are involved. In terms of time, it only utilizes the error nodes of the current time or next time step. More importantly, these error nodes do not need to be calculated recursively in time, but exist in the calculation graph during inference, and can thus realize asynchronous calculation. Compared to the BPTT algorithm in Fig.~\ref{fig:bp}, PCTT algorithm does not require a separate feedback network in both spatial dimension and time domain. 
% This local and asynchronous property of this algorithm may engender a promising prospect of more efficient implementations on neuromorphic hardware and may also facilitate the development of fully distributed neuromorphic architectures. Moreover, it also provides a new idea for solving the problem of gradient explosion/vanishing in BPTT.\
{The inherent locality and asynchronicity of this algorithm hold significant potential for enabling more efficient implementations on neuromorphic hardware and promoting the development of fully distributed neuromorphic architectures. Additionally, it presents a fresh perspective for tackling the gradient explosion/vanishing problem in BPTT.}

\section{Experiments}

\subsection{Experimental Setup }
\subsubsection{Datasets}

We selected four datasets for our experiments: Caltech 101, MNIST, N-MNIST and CIFAR-10 for object classification tasks, and an event camera-based dataset for angular velocity regression.
The Caltech101 Face/Motorbike Dataset is a publicly available classification dataset that includes various images of faces and motorbikes. 
The MNIST dataset is a common benchmark for image classification tasks, comprising 60,000 training images and 10,000 test images. Each image displays a handwritten digit ranging from 0 to 9 and has dimensions of $28\times 28$ pixels. 
he N-MNIST dataset \cite{Orchard2015ConvertingSI} is the spiking version of MNIST dataset which was collected by mounting the ATIS sensor on a motorized translation unit and moving the sensor while viewing the MNIST example on the LCD.  Each dataset sample is 300 ms long and 34×34 pixels big. Different from MNIST data shape $(X , Y)$, the data shape of each input spike of N-MNIST is $(P, X, Y, T)$, where $P$ represents the polarity of the event (positive when the pixel brightness increases, negative when the brightness decreases); $X=Y=34$ is the visual scale.
The CIFAR-10 dataset \citep{krizhevsky2009learning} is a collection of 60,000 32x32 color images in 10 classes, with 6,000 images per class. The Event Camera Angular Velocity Dataset is an open-source synthetic dataset \citep{gehrig2020event}, which uses ESIM \citep{rebecq2018esim} as the event camera simulator. This dataset matches DAVIS240C Event-camera with a resolution of $240\times180$ and selects 10000 panoramic images from a sub-set of Sun360 dataset \citep{xiao2012recognizing}. The random rotational motion used to generate this data covers all axes evenly, resulting in uncorrelated angular velocities across the entire dataset with a mean value of zero. 

\subsubsection{Implementation Details}
For the Caltech101 Face/Motorbike dataset, we implemented a PC-SNN with a network architecture of 28$\times$50-200-2. The output layer consists of two Integrate-and-Fire (IF) neurons, each representing either a face or a motorbike. We initialized the weights of both the hidden and output layers randomly from uniform distributions within the ranges [0, 1] and [0, 5], respectively.
For the MNIST dataset, we employed a PC-SNN with a size of $28\times 28-200-10$ that used first-spike temporal coding in the input layer with maximum simulation time $t_{max}=256$. We varied the variance of the hidden and output layers in our probabilistic model $(\Sigma^{(1)}=10$, $\Sigma^{(2)}=20)$ and set learning rates for the hidden and output layers at $\eta=0.06$ and $\eta=0.02$ respectively. The threshold for membrane potential was set at 100, while the distance term in target firing time was $\gamma=20$. Synapse weights were randomly initialized from uniform distributions ranging between $[0,5]$ and $[0,10]$ for the two different layers. 
For the N-MNIST dataset, we set time interval $T$ to 256ms, and employed a PC-SNN with a size of $34\times 34\times 2-500-10$ and $34\times 34\times 2-500-500-10$ respectively. The other parameters were set as shown in Table.~\ref{parameters1}.
To further demonstrate the effectiveness of our method on a large-scale classification dataset, We evaluated our method on CIFAR-10 based on the VGG-11 and VGG-16 network structure, and the other parameters align with MNIST dataset, all details are shown in Table.~\ref{parameters1}.
For the angular velocity regression task, the spiking convolutional neural network is composed of five convolutional layers, a spiking global pooling layer, and a fully connected layer, which outputs the predicted angular velocity values in three dimensions: tilt, pan, and roll. The first four convolutional layers use a step size of 2 to perform spatial downsampling. Starting with 16 channels in the first convolutional layer, the number of channels in each subsequent convolutional layer is doubled. Shallow feature extraction module parameters remain frozen, while fully connected layer parameters are updated. To balance computation time and performance, the number of inferences was set to 5. In our simulation, we set $T_{0} = 0$ms, $T_{1} = 100$ms, $dt = 1$ms, and the rest parameters were set in Table~\ref{parameters2}.

Moreover, the training for the PC-SNN classification model runs for over 100 epochs and is manually stopped, and each sample undergoes 10 inference iterations during the E-step to ensure stable convergence of the predictive coding inference process. For the ASRNN-PCTT regression model, 5 inference iterations are performed within each 100\,ms time window with 1\,ms time resolution. SGD is adopted as the primary optimization strategy; for PC-SNN, the learning rates are configured as 0.04 and 0.02 for the hidden and output layers, respectively, with a decay factor of 0.5 applied every 10 epochs, while the ASRNN-PCTT model uses a learning rate of $2\times10^{-4}$. For completeness, the Adam optimizer is also implemented as an alternative, using $\beta_{1}=0.9$, $\beta_{2}=0.999$, and $\varepsilon=10^{-8}$. To mitigate overfitting, L2 weight regularization with $\lambda=5\times10^{-6}$ is employed. In addition, a dead neuron reset strategy is applied, wherein neurons that fire in fewer than 0.1\% of training samples per epoch are reinitialized to maintain sufficient network activity and expressive capacity throughout training.

\begin{table}[hbt!]
  \centering
  \caption{Model parameters setting for PC-SNN}
  \begin{tabular}{*{1}{l}|*{6}{c}}
  \toprule
      \multirow{2}{*}{Dataset} & \multicolumn{6}{c}{Model Parameters}   \\ 
      & \multicolumn{1}{l}{$t_{max}$} & \multicolumn{1}{l}{$\vartheta$} & \multicolumn{1}{l}{$\gamma$} & \multicolumn{1}{l}{$\alpha$} & $\eta$ & \multicolumn{1}{l}{$\Sigma$} \\ 
    \midrule
    Caltech & 256 & 100  & 8 & 1 & 0.1  & 10 \\ 
    MNIST & 256 & 100 & 20 & 1 & [0.06,0.02] & 10 \\ 
    N-MNIST & 256 & 100 & 20 & 1 & 0.02 & 5 \\ 
    CIFAR-10 & 256 & 100 & 20 & 1 & [0.1,0.06,0.02] & 10 \\ 
  \bottomrule
  \end{tabular}
  \label{parameters1}
\end{table}

% \subsubsection{Event Camera Angular Velocity Dataset}
% For evaluating our predictive coding learning method in regression task, we employ an open-source synthetic dataset \citep{gehrig2020event}, which uses ESIM \citep{rebecq2018esim} as the event camera simulator. ESIM draws the image along the trajectory and produces an approximation of the brightness of each pixel by interpolation of the brightness signal.This signal is then compared to an artificially selected threshold and used to generate events.This dataset matches DAVIS240C Event-camera with a resolution of $240\times180$.This dataset selects 10000 panoramic images in a sub-set of Sun360 dataset \citep{xiao2012recognizing}. The random rotational motion used to generate this data is to cover all axes evenly. In this way, the angular velocities are uncorrelated over the entire data set, and the mean value is zero. In the experiment, we choose to limit the simulation interval to 100 ms and time step to 1 ms. The training set is divided into 10 groups of data covering the range of angular velocity from low to high.

\begin{table}[hbt!]
    \centering
  \caption{Model parameters setting for ASRNN}
  \begin{tabular}{*{1}{l}|*{7}{c}}
  \toprule
  \multirow{2}{*}{Method} & \multicolumn{7}{c}{Model Parameters}     \\ 
                            & \multicolumn{1}{l}{$b_{0}$} & \multicolumn{1}{l}{$R_{m}$} & $\beta$ & \multicolumn{1}{l}{$\tau_{m}$} & \multicolumn{1}{l}{$\tau_{adp}$} & \multicolumn{1}{l}{$\Sigma$} &\multicolumn{1}{l}{$lr$}\\ 
                            \midrule
  ASRNN-based BPTT & 0.1                            & 6   & 2.0 &4      &700   &-   &2e-3\\
  ASRNN-based PCTT    & 0.1                            & 6   & 2.0 &4      &700   &10   &2e-4\\ \bottomrule
  
  \end{tabular}
  \label{parameters2}
\end{table}

\subsection{Evaluation Metrics}
To measure the classification performance of SNNs, we employ the following accuracy metric: the percentage of correctly classified instances by the model. For the quantitative evaluation of regression performance, the criteria used in the event camera angular velocity prediction experiment include the following two measures:

(1) Root mean Square Error (RMSE)
\begin{equation}\label{RMSE}
    RMSE = \frac{1}{T_{1}-T_{0}} \int_{T_{0}}^{T_{1}} \sqrt{\boldsymbol{e}(t)^{\top} \boldsymbol{e}(t)} \mathrm{d} t
\end{equation}

(2) Relative error
\begin{equation}\label{Relative error}
    R_e  = \frac{\|\boldsymbol{\omega}(\mathrm{t})-\bar{\boldsymbol{\omega}}(t)\|}{\|\bar{\boldsymbol{\omega}}(t)\|}
\end{equation}

\begin{table*}[h]
  \centering
  % \color{blue}
  \caption{Comparison of the Classification Accuracy of Different SNN Models on the Caltech face/motorcycle, MNIST, N-MNIST and CIFAR-10 datasets}
  \label{table:classification_result}
    \resizebox{\textwidth}{!}{
  \begin{tabular}{ccccccc}
  \hline
  \hline
  Dataset & Method & Learning rule  & Network Architecture &  Coding & Neuron model & Acc. (\%)  \\
  \hline
  &R-STDP \citep{stdp_mozafari2018first} & Reward modulated STDP & - & Temporal & Rectified linear   & 98.2 \\
  &SDNN  \citep{kheradpisheh2018stdp}  & unsupervised STDP  & - & Temporal & LIF    & 99.1 \\
  &S4NN  \citep{kheradpisheh2020S4NN}   & Temporal Backprop & - & Temporal & IF    & 99.2 \\
  Caltech face/motorcycle &STiDi-BP \citep{mirsadeghi2021stidi} & Spike time displacement BP& - & Temporal & linear SRM   & 99.2 \\
  &SSTDP \citep{liu2021sstdp}  & SSTDP & - & Temporal & IF    & 99.3  \\
  & \textbf{PC-SNN(Ours)} & predictive coding & - & Temporal & IF    & 99.3 \\
  \hline
  &BP-STDP \citep{tavanaei2019bp} & Backprop using STDP & 784FC-1000FC-10FC & Rate & IF    & 96.6 \\
  &S4NN \citep{kheradpisheh2020S4NN}  & Temporal Backprop & 784FC-400FC-10FC & Temporal & IF    & 97.4 \\
  MNIST  &STiDi-BP \citep{mirsadeghi2021stidi} & Spike time displacement BP& 40C5-P2-1000FC-10FC & Temporal   & linear SRM & 99.2 \\
  &SSTDP \citep{liu2021sstdp}  & SSTDP & 784FC-300FC-10FC & Temporal & IF    & 98.1  \\
  & \textbf{PC-SNN(Ours)} & Predictive coding&  784FC-200FC-10FC  & Temporal  & IF    & 98.1 \\
  \hline
  &SPA \citep{Liu2020EffectiveAO} & Probability-maximization& HMAX-S1-C1-FC   & Rate & LIF   & 96.3 \\
  &lee et al. \citep{Lee2016TrainingDS} &Spike-based Backprop  & 34*34*2-800-10   & Rate & LIF   & 98.6 \\
  N-MNIST&SLAYER \citep{Shrestha2018SLAYERSL} & Spike-based Backprop& 34*34*2-500-500-10  & Rate & SRM   & 98.8 \\
  & \textbf{PC-SNN(Ours)} & Predictive coding &  34*34*2-500-10   & Temporal  & IF    & 97.5\\
  & \textbf{PC-SNN(Ours)} & Predictive coding &  34*34*2-500-500-10   & Temporal  & IF    & 98.5 \\
  \hline
  & SPIKE-NORM \citep{sengupta2019going} & Surrogate BP & VGG-16 & Rate & IF    & 91.55 \\
  & PTL \citep{wu2021progressive} & Surrogate BP & VGG-11 & Rate & IF    & 91.24 \\
  & T2FSNN \citep{park2020t2fsnn} & surrogate gradient & VGG-16 & Temporal & LIF    & 91.43 \\
  & SSTDP \citep{liu2021sstdp} & SSTDP & VGG-7 & Temporal & IF    & 91.31 \\
  CIFAR10 & TSC \citep{ann2snn_han2020deep} & surrogate gradient & VGG-16 & Temporal & IF    & 93.63 \\
  & LC-TTFC \citep{yang2023lc} & Surrogate BP & VGG-11 & Temporal & ReL-PSP & 91.25 \\
  & LC-TTFC \citep{yang2023lc} & Surrogate BP & VGG-16 & Temporal & ReL-PSP & 92.72 \\
  & Spiking-ResNet-34 \citep{Zheng2020GoingDW} & Spike-based BP & ResNet-34 & Rate & ReL-PSP & 93.6 \\
  & \textbf{PC-SNN(Ours)} & predictive coding & VGG-11 & Temporal & IF    & 91.29 \\
  & \textbf{PC-SNN(Ours)} & predictive coding & VGG-16 & Temporal & IF    & 92.51 \\
  \hline
  \hline
  \end{tabular}
    }
\end{table*}

\subsection{Experimental Results}
\subsubsection{Classification}
For classification task, we run our PC-SNN algorithm on two different public image classification datasets to evaluate the local compute predictive coding block. Experiments and analysis are performed to compare our method with other non-convoluted spike-based method. Our experiment goal is to demonstrate the predictive coding block is able to achieve comparable performance levels and simultaneously provides the advantages of biological plausibility and local computation in the same time. 
% The proposed model is based on a straightforward architecture consisting of two fully connected layers in a PC-SNN network. This architecture can be analogized to a simple two-layer MLP, featuring a modest number of parameters and a very straightforward network structure (H*W*2-500-500-N).
We evaluated our PC-SNN method on the Caltech face/motorcycle dataset. Detailed results are shown in Table.~\ref{table:classification_result}. Our PC-SNN method achieved 99.3\% top-1 accuracy, surpassing existing SNN method. Second, we evaluated our method on the MNIST and NMNIST dataset, where the PC-SNN achieved competitive result of 98.1\% and 98.5\%, compared to other SNN methods. Finally, we evaluated our method on the CIFAR-10 dataset, where the PC-SNN achieved comparable performance of 92.51\%, compared to other SNN methods. Details are shown in Table.~\ref{table:classification_result}.

\begin{figure*}[hbt!]
  \centering
  \includegraphics[width=1.0\linewidth]{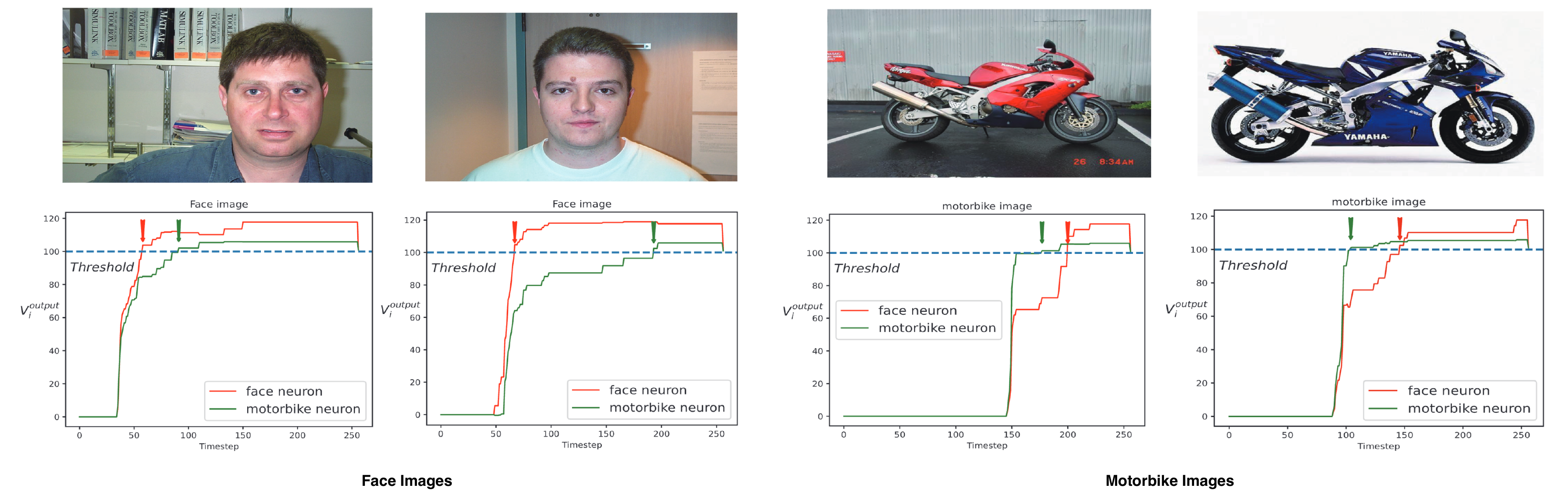}
  \caption{{The temporal dynamics of membrane potentials for output neurons were recorded in response to a subset of face and motorbike images from the Caltech101 dataset. Arrows denote the precise timing of action potentials for each corresponding neuron.}}
  \label{fig:example}
\end{figure*}

% For verify and analyze the functionalities of the proposed PC-SNN, we investigated the membrane potential of output neurons, i.e., $V^{o}_{i}$ after training. As depicted in Fig.~\ref{fig:example}, we present the classification results of four sample colored images with the proposed PC-SNN. 
% Each of the colored images was first resized into $28\times 50$ pixels and converted into grayscale before encoding using TTFS coding strategy. Then, with the pixel density of grayscale and $28\times 50$ input neurons, the TTFS technique encodes each image into binary spike train consisting of only zeros and ones. In this coding scheme, the initial spike activation time of each neuron is determined by the magnitude of the pixel density. Greater pixel values correspond to earlier emission times. After integration of the spike trains of the hidden layer, the corresponding two output neurons' membrane potentials are in the time interval of $[0,t_{max}]$. 

{To verify and analyze the operational characteristics of the proposed PC-SNN, the membrane potential of the output neurons, denoted as $V^{o}_{i}$, was investigated following the training phase. As illustrated in Fig.~\ref{fig:example}, the classification results for four sample color images using the proposed PC-SNN are presented. Each color image initially underwent preprocessing, which involved being resized to $28\times 50$ pixels and converted to its grayscale equivalent, before being encoded using the TTFS methodology. Subsequently, based on the grayscale pixel intensities across the $28\times 50$ input neuron array, the TTFS technique encodes each image into a binary spike train. In this encoding scheme, the initial spike timing of each neuron is contingent upon the magnitude of the pixel intensity, where greater pixel values correspond to earlier spike emissions. Upon the integration of spike trains from the hidden layer, the membrane potentials of the respective output neurons are recorded within the time interval $[0, t_{max}]$.}

\begin{figure}[hbt!]
  \centering
  \includegraphics[width=0.7\linewidth]{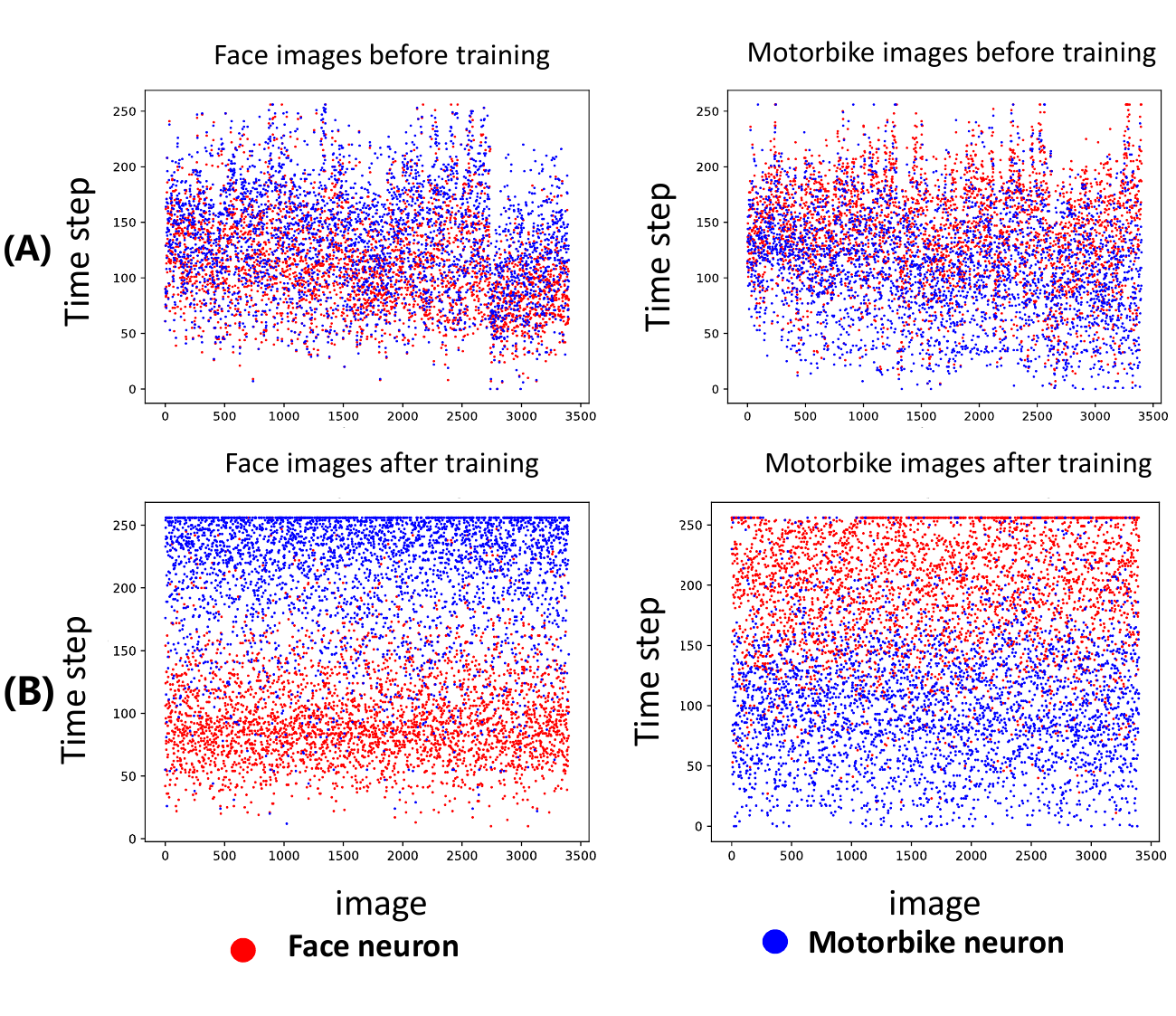}
  \caption{{This figure presents a comparative analysis of the firing time distributions for two output neurons in response to face and motorbike stimuli. The data are depicted for two distinct conditions: (A) prior to training and (B) subsequent to training.}}
  \label{fig:scatter}
\end{figure}

\begin{figure}[hbt!]
  \centering
  \includegraphics[width=0.6\linewidth]{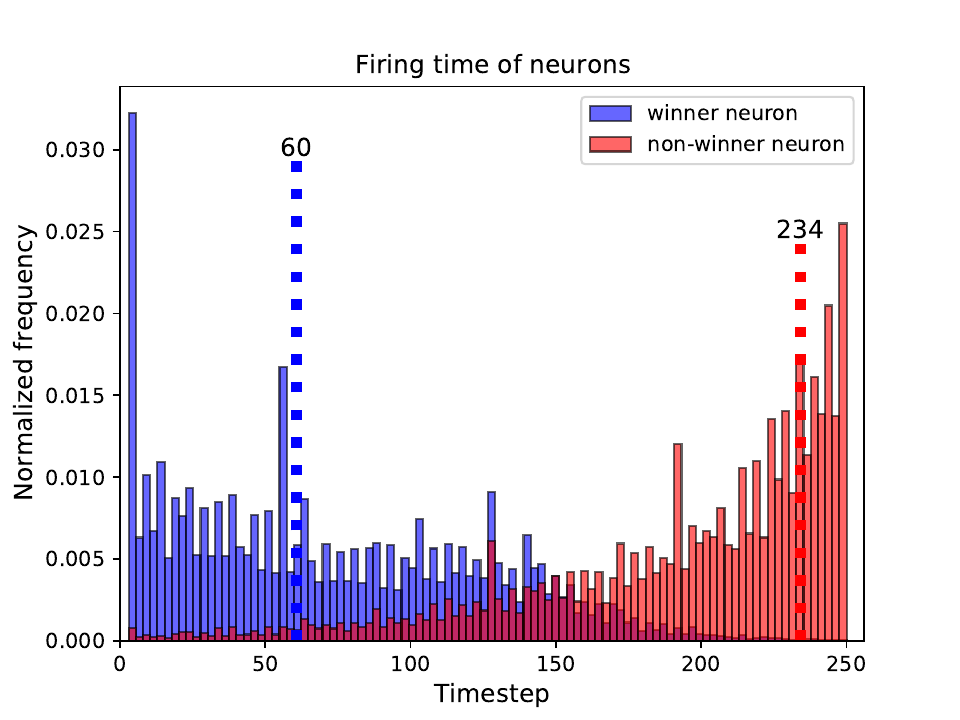}
  \caption{{This figure presents a histogram comparing the firing time distributions for two distinct neuronal populations in response to the training dataset. The blue and red bars represent the winner and non-winner neurons, respectively. Additionally, the mean firing time for each respective neuron group is denoted by a dashed line.}}
  \label{fig:histogram}
  \end{figure}

\begin{figure}[hbt!]
  \centering
  \includegraphics[width=0.6\linewidth]{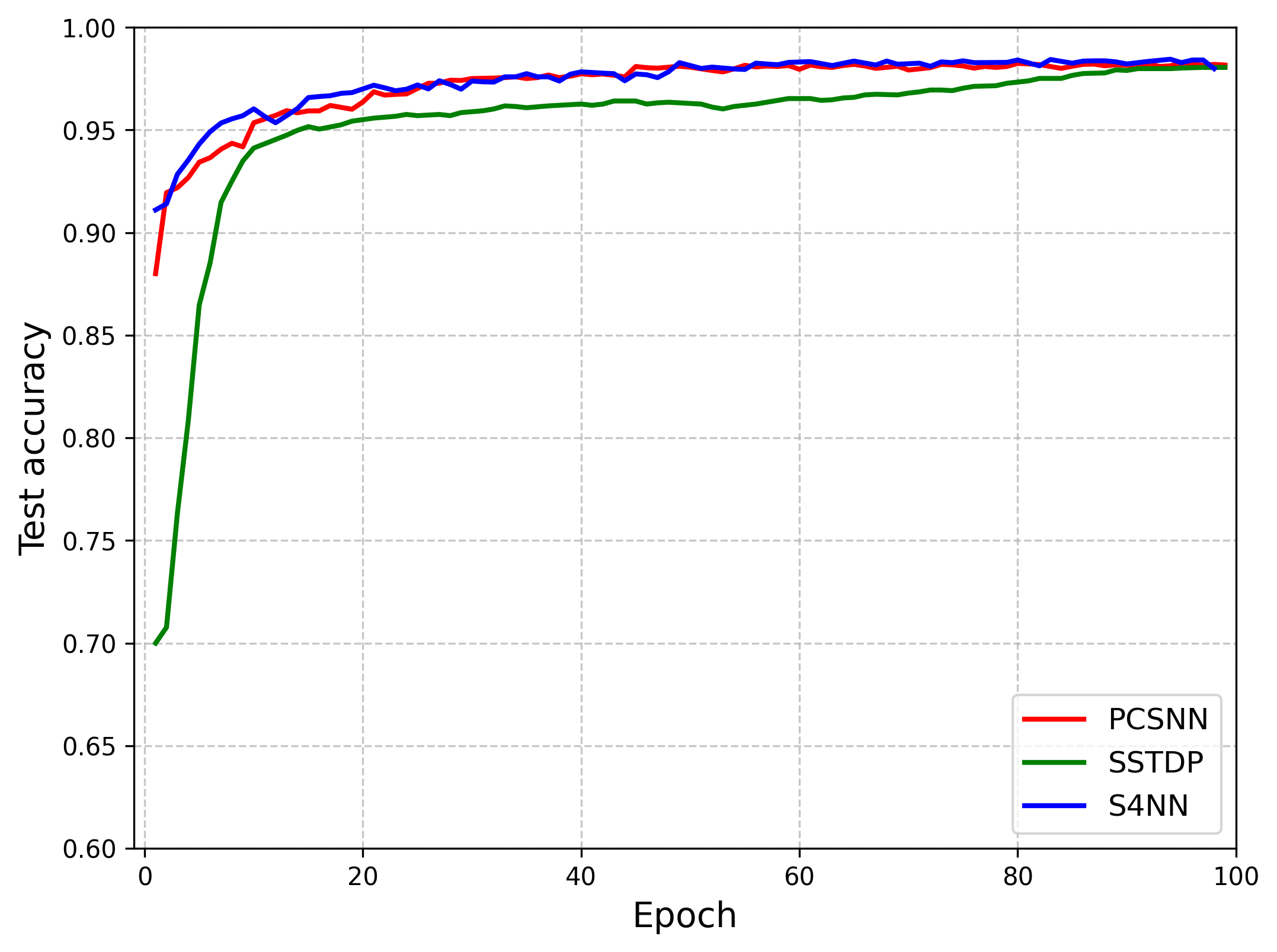}
  \caption{Comparison of MNIST test accuracy convergence over 100 epochs for three SNN models: PCSNN, SSTDP, and S4NN.} 
  \label{fig:converge}
\end{figure}

% For every sample image, the target neuron's membrane potential grows faster and reaches the threshold earlier than that of another neuron. In Fig.~\ref{fig:scatter}, we investigated the firing time distribution of the face and motorbike output neurons over 3600 face and 3600 motorbike images at A) the beginning and B) the end of the learning process. As we can see, the firing time of the two output neurons before training was disordered, indicating that the spiking neuron network was not capable of classification at this time. After training, the neuron corresponding to the correct category fired significantly earlier than the other. 
{Upon presentation of a sample image, the membrane potential of the target-specific neuron is designed to increase at a faster rate, thereby reaching the firing threshold with a lower latency compared to other neurons. An investigation into the firing time distributions of the face- and motorbike-selective output neurons is presented in Fig.~\ref{fig:scatter}. The analysis utilized a dataset comprising 3600 face and 3600 motorbike images and was conducted at two distinct stages: (A) before the commencement of training and (B) after its conclusion. The initial, pre-training results indicate a disordered firing pattern between the two output neurons, demonstrating the spiking neural network's initial inability to perform the classification task. Conversely, after the training process, the neuron corresponding to the correct image category exhibited a significantly earlier firing time than its counterpart.}
Some neurons gather at $t_{max}$ because if they are not activated within a set time interval, their firing time is artificially set to $t_{max}$. Fig.~\ref{fig:histogram} shows the firing time distribution of the winner neuron (earliest activated neuron) and non-winner neurons. About 90\% of winner neurons activate before 150 timesteps, with an average firing time of 60 timesteps. In contrast, 85\% of non-winner neurons fire after 150 timesteps, with an average of 234 timesteps. This clear separation aids network training and judgment, with PC-SNN responding to input images within about 60 timesteps 50\% of the time. Fig.~\ref{fig:converge} shows the convergence speed of the PC-SNN, SSTDP \citep{liu2021sstdp}, and S4NN  \citep{kheradpisheh2020S4NN} on the MNIST dataset. As demonstrated, the PC-SNN converges faster than the SSTDP and S4NN.  
he linear approximation constant $\alpha$ represents the rising slope of membrane potential 
near the firing threshold. From Equations (20)-(22), we observe that the effective learning rate $\eta_{\text{eff}} \approx \eta/\alpha$, indicating a coupling between $\alpha$ and $\eta$. We conducted five sets of comparative experiments using \( \alpha = [0.1, 0.5, 1, 5, 10] \) while keeping learning rate $\eta$ fixed. The results show that the model exhibited near non-convergence for \( a = 0.1, 10 \). For \( a = 0.5, 5 \), the convergence was significantly slow, and the accuracy failed to exceed 90\% even after training for more than 100 epochs. Moreover, The parameter $\gamma$ defines a temporal classification margin. It controls the minimum separation between correct and incorrect class firing times, improving robustness against neural noise. Four groups of controlled experiments are conducted with $\gamma$ values set to 5, 10, 20, and 40, respectively. Experimental results indicate that the convergence behaviors for $\gamma$ values of 10 and 20 are similar. In contrast, when $\gamma$ is set to 5 or 40, convergence is markedly slower and fails to reach the optimal solution.

\begin{figure}[tbp]
  \centering
  \includegraphics[width=0.6\linewidth]{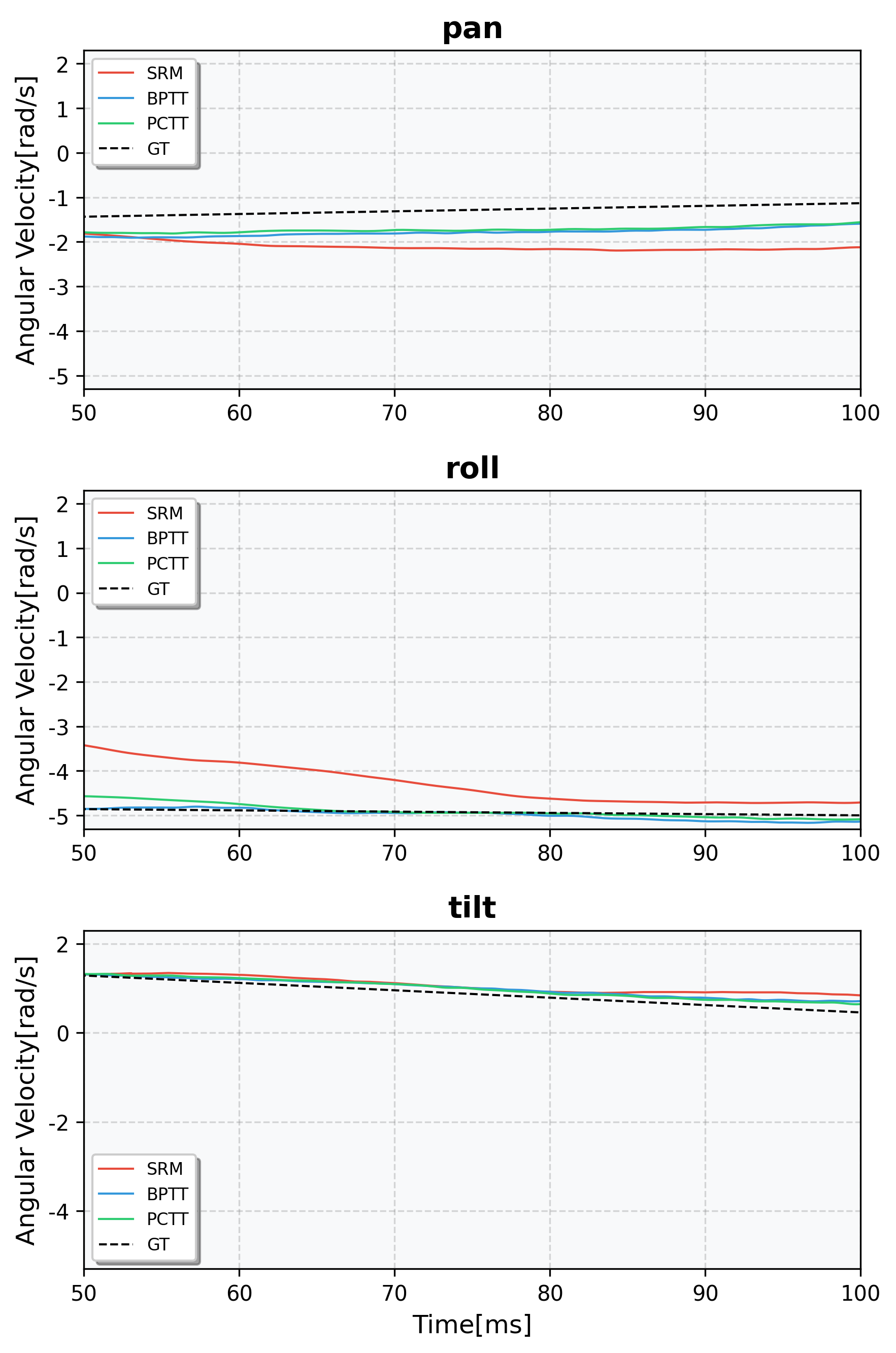}
  \caption{Continuous-time angular velocity predictions by three algorithms (SRM, BPTT, PCTT) and the corresponding ground truth.}
  \label{fig:4samples}
\end{figure}

\begin{figure}[tbp]
  \centering
  \includegraphics[width=0.5\linewidth]{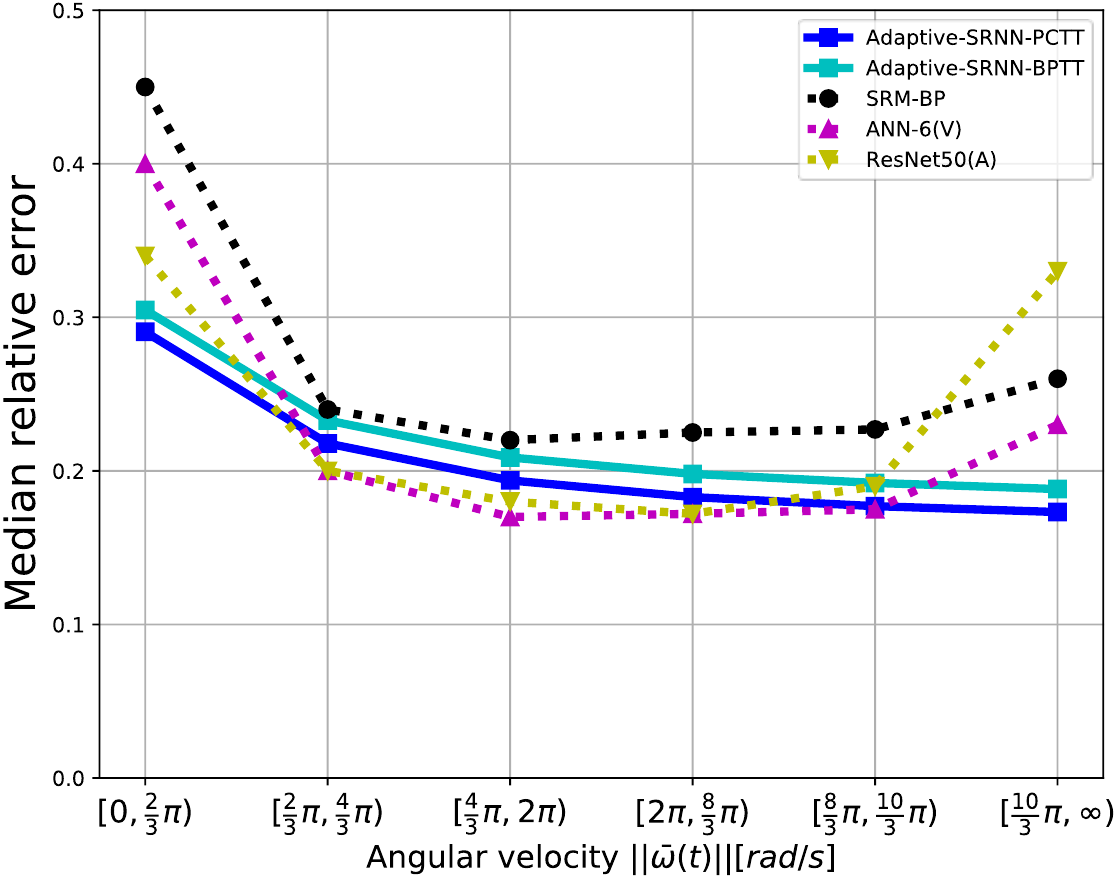}
  \caption{Median relative Error of all baseline models in different angular velocity ranges of the test set}
  \label{fig:median_relative_error}
\end{figure}

\subsubsection{Regression}
For the regression task, we explored the applicability of ASRNN to predict angular velocity of a rotating event camera. We compared existing methods to our ASRNN method trained using either BPTT or PCTT. The performance comparison of several baseline models are lised in Table~\ref{tab:angular}.
The spiking convolutional network structure of the SRM-BP algorithm \citep{gehrig2020event} is the same  in this work. 
ANN-6 is a 6-layer convolutional neural network using ReLU as the activation function, and "V" indicates Voxel-Grid events. The input event type used by the ResNet-50 algorithm is "A," which represents the two-channel frames calculated through accumulating events, while "E" represents the original events from the event cameras. The prediction gap between the baseline obtained by the ResNet-50 and ANN-6 architectures is due to differences in their input representations. Unlike the Voxel-based method, the Accumulation-based method discards the time of the event completely. This significantly influences the performance of continuous-time regression prediction of angular velocity.

\begin{table}[h]
  \centering
  \caption{Event Camera Angular Velocity}
  \label{tab:angular}
  \resizebox{0.7\textwidth}{!}{
  \begin{threeparttable}
  \large
  \begin{tabular}{*{1}{l}|*{3}{c}}
      \midrule
      Method & Input Type &  Relative error & RMSE (deg/s)  \\
      \midrule
      ANN-6       & V  &  0.22& 59.0  \\
      ResNet-50  & A   & 0.22 & 66.8 \\
      SRM-BP \citep{gehrig2020event}     & E & 0.26  & 66.3 \\
      ASRNN-BPTT   & E & 0.22 & 60.39 \\
      ASRNN-PCTT     & E &   0.22 & 59.58 \\
      \bottomrule
  \end{tabular}
  \begin{tablenotes}
    \footnotesize
    \item V indicates that Voxel-Grid event
    \item A indicates two-channel frames calculated through accumulating events
    \item E indicates the original events
  \end{tablenotes}
\end{threeparttable}
  }
\end{table}

ASRNN-Based BPTT Algorithm proposed in this paper achieves better performance compared with the previous work -- SRM-based backpropagation algorithm: Our relative error is 0.22 and RMSE is 60.39 deg/s, which improves the RMSE by 6 deg/s compared with the previous SRM model, and is better than the Accumulation-based ResNet-50 algorithm. At the same time, this result is comparable to the result of Voxel-Based ANN-6 algorithm. Moreover, in our method, the event-based input does not need any preprocessing, it really realizes the end-to-end angular velocity prediction. The RMSE obtained by the PCTT algorithm is 59.58 deg/s and Relative error is 0.22, which is closer to the current ANN best result of 59.0 deg/s. This may be attributed to the elimination of iteration in time, solving the possible problem of gradient explosion or vanishing in BPTT. In addition, PCTT has the advantage of biological feasibility and satisfies the local plasticity in biological neurology.

We illustrate the 50 - 100ms continuous-time angular velocity prediction of random sample of ASRNN-based BPTT algorithm and PCTT algorithm in Fig.~\ref{fig:4samples}. Compared with SRM-based algorithm, it can be concluded from qualitative analysis that the ASRNN-based algorithm can achieve more accurate predictions. The predicted angular velocity in three directions can all well track the ground truth angular velocity, and the error between the predicted value and the ground truth is smaller than SRM-based algorithm.
Fig.~\ref{fig:median_relative_error} and Table~\ref{tab:angular} illustrate the median relative errors of all baseline models. At low angular velocities, all models exhibit relatively high errors (~0.3), with ASRNN-based BPTT (0.31) and PCTT (0.29) outperforming other baselines. At high angular velocities, our proposed methods achieve errors below 0.2. For intermediate velocities, our methods perform comparably to ANN-6 (A) and ResNet-50 (V), with median relative errors of approximately 0.22. Overall, the proposed ASRNN-based approaches demonstrate superior performance at both low and high angular velocity ranges.

% The median relative errors of all the baseline models listed in Table \ref{tab:angular} are shown in Fig.~\ref{fig:median_relative_error}.
% When the angular velocity is low, the relative errors of all models are relatively large, around 0.3. In practice, it is often difficult to achieve low relative errors at low velocities because the relative errors are greater at angular velocities near zero. In the low angular velocity case, ASRNN-based BPTT algorithm achieved 0.31 and PCTT algorithm achieved 0.29, which are better than other baseline models. When the angular velocity is large, our proposed methods are both lower than 0.2. For the angular velocity values in the intermediate range, our proposed methods have the same performance as two ANN baseline models: ANN-6 (A) and ResNet-50 (V), with a median relative error of about 0.22. In general, ASRNN-based BPTT and PCTT algorithms proposed in this work can achieve smaller median relative error in the case of smaller and larger angular velocities. In the middle range of angular velocity, ANN-6 and ResNet-50 (V) perform better.

\subsection{Time Complexity Analysis}
The time complexity of PC-SNN for training K-layer SNNs with biologically plausible algorithms offers the advantage of lower algorithmic complexity. While both PC-SNN and BP-SNN require a forward pass to compute spike times, the key distinction lies in the subsequent learning phase. We denote the computational costs of one-step feedforward and one-step local-plasticity propagation as $n$ and $m$, respectively. 
% The PC-SNN completes the feedforward procedure with a cost of $\mathcal{O}(nK)$ where $K$ is the number of layers, then implements the inference process for all hidden layers in parallel using only local information. After that, it updates the network parameters for all hidden layers in a local-plasticity manner also in parallel. Therefore, the algorithm complexity of PC-SNN is $\mathcal{O}(nK + 2mK)$. 
PC-SNN completes the feedforward step at a cost of $\mathcal{O}(nK)$, where $K$ is the number of layers. It then performs parallel inference across all hidden layers using local information and updates the network parameters for each layer in a local-plasticity manner, also in parallel. Thus, the algorithm complexity is $\mathcal{O}(nK + 2mK)$.
In contrast, BP-SNN not only incurs a feedforward cost of $\mathcal{O}(nK)$ but also involves multistep backpropagation layer by layer using differential chain rule with computing cost of $\mathcal{O}(nK + (m + mK)K)$ \citep{zhang2021tuning}. 
The PC-SNN algorithm updates spiking temporal activities and weights using only information from directly connected nodes, allowing for parallel implementation. 
% This local and parallelizable property makes our algorithm a promising candidate for more efficient implementations on neuromorphic hardware. It may also facilitate the development of fully distributed neuromorphic architectures.
Our algorithm's inherent locality and parallelism make it ideal for efficient neuromorphic hardware implementations and could enable fully distributed architectures.

\subsection{Biological Plausibility}
Learning in the brain occurs through changes in synaptic connections between neurons, and there is very limited evidence suggesting that exact formulations of backpropagation exist in the cortex.   
Despite SNN's biological plausibility nature, primarily learning methods still align with traditional AI,  neglecting its brain-like learning mechanisms and biological plausibility
Our PC-SNN framework relies on a Hebbian-based rule to regulate synaptic plasticity, controlled entirely by local nodes which shares similarities with synaptic plasticity in the human brain \citep{lillicrap2020backpropagation, caucheteuxEvidencePredictiveCoding2023, lillicrapRandomSynapticFeedback2016}.

\section{Conclusion and Discussion}

Our work implement the predictive coding learning framework to spiking neural networks for image and event camera rotation datasets, using only local information and end-to-end training. The results are comparable to state-of-the-art SNN learning methods, showcasing the framework's capabilities in SNN training. 
The proposed networks are evaluated on three images datasets and one event camera datasets: Caltech Face/Motorbike, MNIST, CIFAR10 and event camera angular velocity dataset. 
% Our proposed framework offers two prominent features:
% (1) Biological plausibility is achieved by using only local information in both the inference and weight update process. PC-SNN eliminates the need for a separate feedback network, processing information asynchronously and concurrently in a parallel, local manner.
% (2) Our proposed learning rule suits on-chip learning perfectly due to lower wiring layout complexity derived from elimination of feedback network. Although hardware implementation is beyond scope of this paper, we believe that our model provides research potential on hardware implementation in an energy-efficient manner.
{
The proposed framework is distinguished by two primary characteristics.
(1) It exhibits a high degree of biological plausibility, as both the inference and weight modification processes rely exclusively on local information. This approach obviates the requirement for a distinct feedback network, enabling information to be processed asynchronously, concurrently, and in a localized, parallel fashion within the PC-SNN architecture.
(2) The proposed learning rule is exceptionally well-suited for on-chip implementation. This suitability is a direct consequence of the reduced wiring layout complexity afforded by the elimination of the feedback network. 
To further elaborate on these two characteristics, we discuss the theoretical grounding of our approach from both neuroscience and engineering perspectives.
From a neuroscience perspective, according to predictive coding theory \citep{1999predictiveframework}, higher cortical areas continuously generate predictions about expected sensory inputs, which propagate as top-down signals to lower areas. These predictions effectively establish expected firing patterns that lower-level neurons should match. In our framework, the target firing times at the output layer can be interpreted as such top-down predictions derived from learned categorical representations. In reinforcement learning contexts, dopaminergic and other neuromodulatory systems provide global reward signals that modulate synaptic plasticity. The target spike times in supervised learning can be viewed as analogous to reward-prediction signals that guide learning toward desired behavioral outcomes.
And from a neuromorphic hardware deployment perspective, our PC-SNN algorithm is particularly well-suited for deployment on neuromorphic platforms due to its local computation and parallel update characteristics. Intel's Loihi \citep{loihi} and Loihi 2 \citep{loihi2} chips natively support local learning rules and spike-timing-based computation. The local Hebbian updates in PC-SNN can be directly mapped to Loihi's on-chip learning engines without requiring off-chip gradient computation. Platforms such as BrainScaleS-2 \citep{pehle2022brainscales} and DYNAPs \citep{DYNAPs} implement physical neuron and synapse dynamics. The continuous-time nature of our spiking neuron model naturally maps to analog circuits, while the local plasticity rule avoids the need for complex routing of global error signals.
While a physical hardware implementation is beyond the scope of this study, the model demonstrates considerable potential for future development into energy-efficient neuromorphic hardware.
For the future, the current classification evaluation focuses on Caltech Face/Motorbike, MNIST, N-MNIST and CIFAR-10, with CIFAR-10 experiments conducted on VGG-11/16 backbones. While these results validate the effectiveness of predictive-coding-based local Hebbian learning on both shallow and moderately deep architectures, further validation on larger-scale datasets (e.g., CIFAR-100 and ImageNet) is an important next step to comprehensively assess scalability and generalization.

\normalem 
\bibliography{ref}
\bibliographystyle{apalike}

\end{document}